\documentclass[sigconf]{acmart}
\usepackage{url}  
\usepackage{graphicx}  
\usepackage{amsfonts}
\usepackage{amssymb,amsmath}
\usepackage{xspace}
\usepackage{cleveref}
\usepackage{booktabs}
\usepackage{algorithm}
\usepackage{algorithmic}
\usepackage{amsmath,bm,bbm}
\usepackage{multirow}
\usepackage [autostyle, english = american]{csquotes}
\usepackage{array}
\newcolumntype{L}[1]{>{\raggedright\let\newline\\\arraybackslash\hspace{0pt}}m{#1}}
\newcolumntype{C}[1]{>{\centering\let\newline\\\arraybackslash\hspace{0pt}}m{#1}}
\newcolumntype{R}[1]{>{\raggedleft\let\newline\\\arraybackslash\hspace{0pt}}m{#1}}
\usepackage{enumitem}

\usepackage{caption}
\usepackage{subcaption}

\MakeOuterQuote{"}

\DeclareMathOperator*{\argmax}{arg\,max}

\newcommand{\mname}{\texttt{HOLMES}\xspace}
\usepackage{color}

\definecolor{codegreen}{rgb}{0,0.6,0}
\definecolor{codegray}{rgb}{0.5,0.5,0.5}
\definecolor{codepurple}{rgb}{0.58,0,0.82}
\definecolor{backcolor}{rgb}{0.95,0.95,0.92}
\definecolor{purple}{RGB}{128,0,128}
\definecolor{indigo}{RGB}{75,0,130}
\definecolor{royalblue}{RGB}{65,105,225}
\definecolor{navy}{RGB}{0,0,128}

\newif\ifcommenton
\commentontrue
\ifcommenton
\newcommand{\hsd}[1]{\textcolor{blue}{\emph{[Shenda: #1]}}}
\newcommand{\js}[1]{\textcolor{red}{\emph{[JS: #1]}}}
\newcommand{\yx}[1]{\textcolor{purple}{\emph{[YX: #1]}}}
\newcommand{\at}[1]{\textcolor{royalblue}{\emph{[AT: #1]}}}

\newcommand{\al}[1]{\textcolor{brown}{\emph{[AL: #1]}}}
\else
\newcommand{\hsd}[1]{}
\newcommand{\js}[1]{}
\newcommand{\yx}[1]{}
\newcommand{\at}[1]{}
\newcommand{\al}[1]{}
\fi

\AtBeginDocument{%
  \providecommand\BibTeX{{%
    \normalfont B\kern-0.5em{\scshape i\kern-0.25em b}\kern-0.8em\TeX}}}

\copyrightyear{2020} 
\acmYear{2020} 
\setcopyright{acmcopyright}\acmConference[KDD '20]{Proceedings of the 26th ACM SIGKDD Conference on Knowledge Discovery and Data Mining}{August 23--27, 2020}{Virtual Event, CA, USA}
\acmBooktitle{Proceedings of the 26th ACM SIGKDD Conference on Knowledge Discovery and Data Mining (KDD '20), August 23--27, 2020, Virtual Event, CA, USA}
\acmPrice{15.00}
\acmDOI{10.1145/3394486.3403212}
\acmISBN{978-1-4503-7998-4/20/08}

\begin{document}
\fancyhead{}

\title{\mname: Health OnLine Model Ensemble Serving for\\Deep Learning Models in Intensive Care Units}

\author{Shenda~Hong$^{1,*}$, Yanbo~Xu$^{1,*}$, Alind~Khare$^{1,*}$, Satria~Priambada$^{1,*}$, Kevin~Maher$^2$, Alaa~Aljiffry$^2$, Jimeng~Sun$^3$, Alexey~Tumanov$^1$}
\affiliation{$^1$Georgia Institute of Technology, $^2$Childrens Healthcare of Atlanta, $^3$University of Illinois at Urbana-Champaign}
\thanks{* Authors contributed equally to this research. }

\begin{abstract}

Deep learning models have achieved expert-level performance in healthcare with an exclusive focus on training accurate models. 
However, in many clinical environments such as intensive care unit (ICU), 
real-time model serving is equally if not more important than accuracy, because in ICU patient care is simultaneously more urgent and more expensive. Clinical decisions and their timeliness, therefore, directly affect both the patient outcome and the cost of care. 
To make \textit{timely} decisions, we argue the underlying serving system must be latency-aware. To compound the challenge, health analytic applications often require a combination of models instead of a single model, to better specialize individual models for different targets, multi-modal data, different prediction windows, and potentially
personalized predictions
To address these challenges, we propose \mname---an online model ensemble serving framework for healthcare applications. \mname dynamically identifies the best performing set of models to ensemble for highest accuracy, while also satisfying sub-second latency constraints on end-to-end prediction. We demonstrate that \mname is able to navigate the accuracy/latency tradeoff efficiently, compose the ensemble, and serve the model ensemble pipeline, scaling to simultaneously streaming data from 100 patients, each producing waveform data at 250~Hz. \mname outperforms the conventional offline batch-processed inference for the same clinical task in terms of accuracy and latency (by order of magnitude). \mname is tested on risk prediction task on pediatric cardio ICU data with above 95\% prediction accuracy and sub-second latency on 64-bed simulation.

\end{abstract}


\maketitle
\section{Introduction}
\label{sec:intro}

Vast amount of real-time monitoring data such as electrocardiogram (ECG) and electroencephalogram (EEG) are being observed and collected 
especially in intensive care units (ICUs), which provides invaluable input for training deep learning  (DL) models~\cite{xu2018raim,DBLP:journals/corr/LiptonKEW15,rajkomar2018scalable}. 
In fact deep learning models have achieved state-of-the-art performance in many healthcare and medical applications such as Radiology \cite{rajpurkar2017chexnet}, Ophthalmology \cite{google_jama}, and Cardiology \cite{hannun2019cardiologist,hong2020opportunities}. 
Almost all previous works focused on optimizing deep neural networks for prediction accuracy~\cite{xiao2018opportunities}. 
However, in a high stakes environment such as ICUs 
serving these trained machine learning models in \textit{real time}
is an equally important but hitherto largely overlooked requirement.
In general, a computational graph of stateful and stateless components responsible for capturing and performing prediction on multi-modal sensory inputs must be provisioned to perform  {\bf high accuracy} predictions with {\bf low latency} on a {\bf dynamic} stream of multi-patient data.

\begin{figure}[b!]
\centering
\includegraphics[width=.9\linewidth]{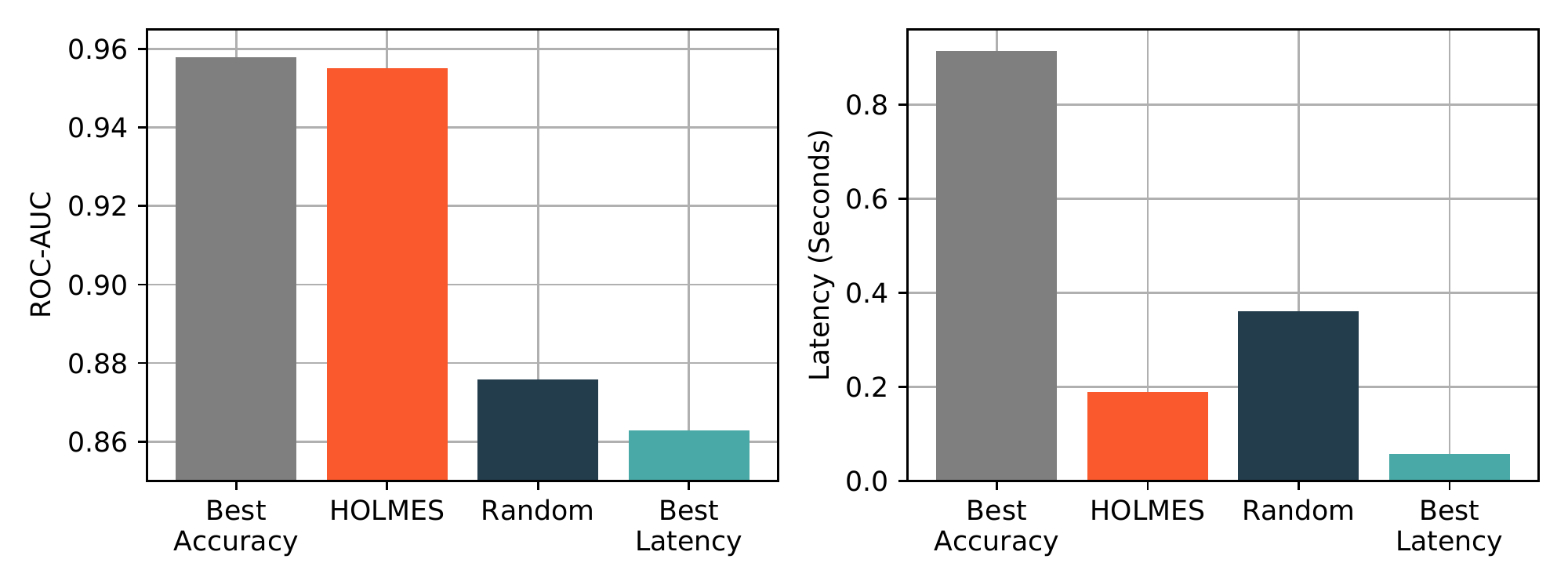}
\vspace{-.5em}
\caption{\mname finds a better balance between accuracy (ROC-AUC) and latency.  \mname reaches competitive accuracy within the 200ms latency budget.}
\label{fig:intro}
\end{figure}

Although recent work on prediction serving systems like TensorFlow Serving~\cite{olston2017tensorflow} and Clipper~\cite{clipper-nsdi17}
 have proposed general frameworks for serving deep models at the system level, they are unable to support the complex needs of health predictive models.
First, these model serving platforms primarily target single model serving and can be thought of as the physical execution layer at the lower level of the stack.
The complexity of health predictive pipelines involves capturing and pre-processing multi-modal sensory data, arriving at different frequencies, and performing periodic inference on this multi-rate, multi-modal stream of data under the constraint of soft real-time latency Service Level Objectives (SLOs). 
Second, in a hospital setting often due to privacy/legal requirements, machine learning (ML) serving systems operate under the constraints of a fixed set of limited on-premises resources. This necessitates a careful exploration of the accuracy/latency tradeoff, whereby higher accuracy can be achieved at the expense of prohibitively high latency. Thus, the soft real-time serving platform needs to be able to explore this tradeoff and navigate it to find the right pipeline configuration for a specific clinical use case at hand.
For example, in ICU environments, a large number of DL predictive models can be constructed based on different data modalities (e.g., ECG, heart rate, blood pressure),  different prediction windows (e.g., 30 sec, 10 min, 1 hour, 1 day) and different patient data (e.g., models are retrained every week based on data from new patients).

\noindent{\bf  Application 1: Length of Stay prediction (Accuracy first)} ICU beds are limited and expensive resources to be managed effectively. 
It is estimated that there are 14.98 beds on average at each ICU \footnote{Using data from American Hospital Association (AHA) statistics. Healthcare Cost Report Information System (HCRIS) data is unavailable.} \cite{halpern2015critical}, median daily cost of ICU ranges from \$6,318 (Medical Surgical ICU) to \$10,382 (Cardiac Surgical ICU) \footnote{Data from Montefiore Medical Center in the Bronx, New York during 2013.} \cite{gershengorn2015patterns}. Accurate real-time prediction of length of stay and readiness for discharge of a patient is an important capability for ICU resource optimization. Model accuracy is essential as a wrong prediction will lead to inadequate resource allocation and sub-optimal care to critical patients. 

\noindent{\bf Application 2: Mortality prediction (Latency first)} ICU patients are vulnerable and unstable. Patient conditions may change quickly over time. Rapid response is the key to save a life. For example, accurate predicting cardiac arrest will happen even in the next 10 minutes can save lives. In this rapidly changing situation, the latency of the prediction is crucial. We want to be able to utilize most recently patient data to perform the prediction. 

\begin{figure}
\centering
\includegraphics[width=0.7\linewidth]{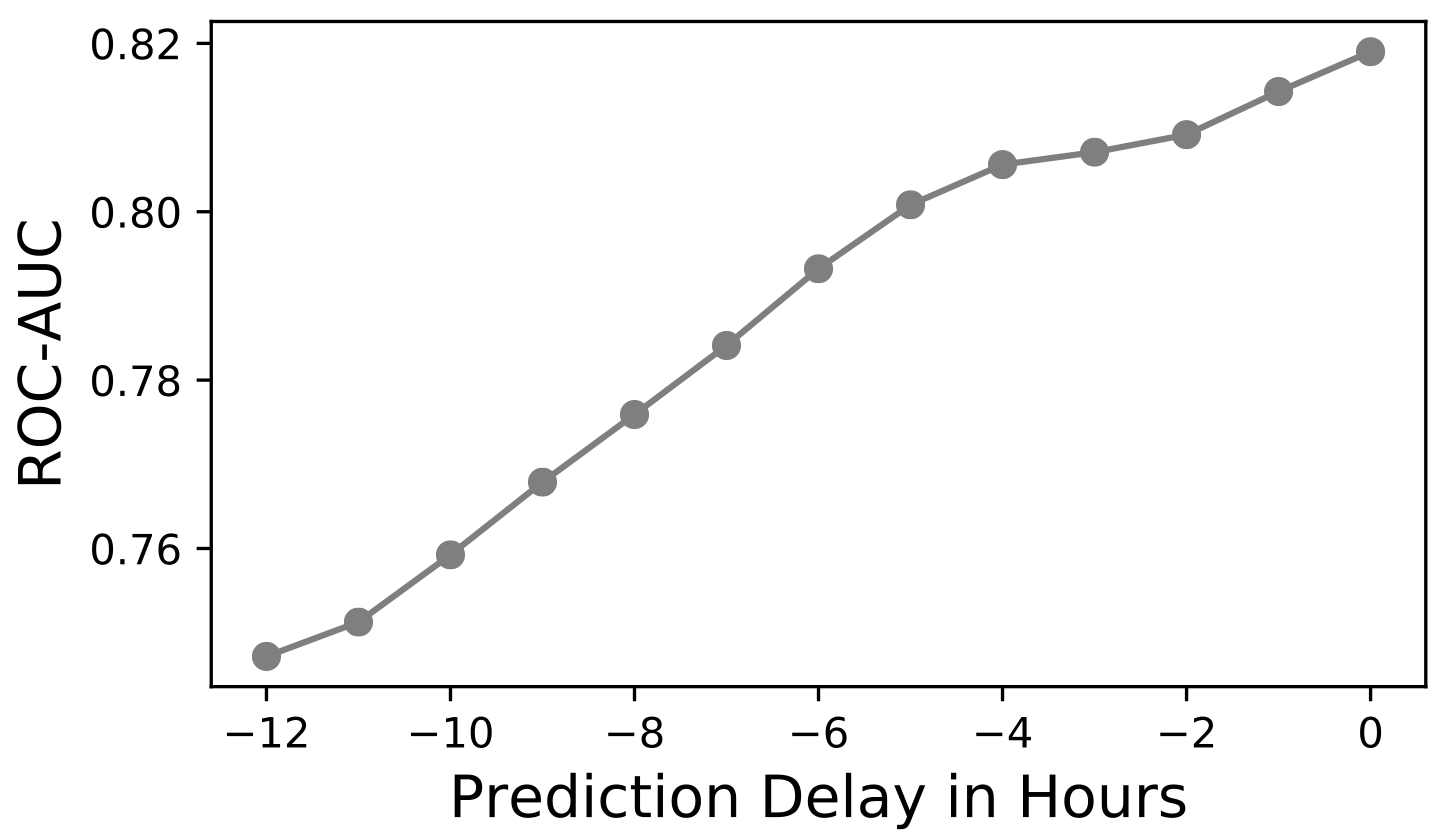}
\vspace{-.5em}
    \caption{
        Accuracy decreases with prediction delay. 
        Prediction accuracy plotted for CICU patient's readiness for stepdown transfer based on 3-lead continuous ECG waveforms.   
    }
\vspace{-1.5em}
\label{fig:intro2}
\end{figure}

Deep learning models for ICU environments have drawn much more attention including risk prediction \cite{xu2018raim} and treatment recommendation~\cite{wang2018supervised}. However, it is difficult to support real-time model serving in ICU due to the following challenges: 
\begin{itemize}[leftmargin=5mm]
\itemsep0em
\item {\bf Multiple data modalities:} Many different data modalities are captured in ICUs.  Some modalities like Electrocardiogram (ECG) signals have high sampling frequency up to 1,000 Hz, while other modalities like medication only occurs once a few hours. 
It is unnecessary and nearly impossible to train and serve a unified model that serves all data modalities.
\item {\bf Noisy environment:} ICUs are highly dynamic environments. Data collections can be unreliable especially for sicker patients as more interventions can happen to them which often affect the monitoring device (e.g., sensors fall off or are removed during interventions)
\item \textbf{Heterogeneous conditions:} Diverse patients with very different conditions are present in ICU, which leads to different risk and monitoring needs. 
For example, complex patients after surgery may need to be monitored and assessed more frequently by multiple models, compared to stable patients who are close to being moved out of the ICU. 
\item \textbf{Limited computational resources:} Hospital is a typical resource constrained environment including computing. Most of the bedside computing or local clusters are limited in computing capability (e.g., the limited number of GPUs and memory). 
Real-time data capturing and model serving workload can be overwhelming for that hardware. While recent works can achieve predictive accuracy in ICU applications, those models can often be too heavy and inflexible to deploy. Since conditions of ICU patients often change quickly, a small latency for model serving is required especially for latency first tasks like cardiac arrest prediction.
\end{itemize}

We propose \mname---a novel real-time model ensemble composition and serving system for deep learning models with clinical applications in the ICU with the following contributions: 
\begin{itemize}[leftmargin=5mm]
\itemsep0em
    \item {\bf Model zoo:} Instead of the common approach to train one unified model serving all patients with the same set of data modalities, \mname trains a set of models (called {\it model zoo}). Each model can be specialized for one particular task using one data modality with a certain, independently chosen time window. 
\item {\bf Ensemble composer:} \mname leverages sequential model-based Bayesian optimization with genetic exploration to select a subset of models from the model zoo for serving different patients under a restricted set of computational resources and within a specified latency budget. 
\item {\bf Real-time model serving:} \mname extends the Ray~\cite{moritz2018ray} framework with a real-time model pipeline serving functionality. 
\item {\bf Real-data experiment:} We extensively evaluate \mname on real ICU data in a realistic simulated streaming environment. Our experiments confirm \mname can simultaneously serve up to 100 beds of streaming data while achieving 95\% accuracy for stepdown readiness prediction with sub-second latency. 
\end{itemize}

In summary, \mname' combination of the model ensemble search algorithm and a latency-aware, soft real-time model pipeline serving framework enables exploration of the accuracy/latency tradeoff space for a variety of relevant clinical use cases.

\section{Related Work}
\label{sec:related}

\mname draws its benefits from a combination of the automatic model ensemble composition algorithm and the underlying
latency-sensitive model pipeline serving framework. In this section, we show how \mname builds on and compares to the state-of-the-art for both. The key takeaway in this comparison is two-fold. First, \mname proposes automatic model ensemble construction from specialized models that can individually be more narrowly trained to specialize in different data modalities and observation windows. This is a departure from a body of literature that depends on a single model. Second, \mname builds on a latency-aware system that is simultaneously capable of serving clinical use cases across the spectrum of prediction latency requirements. Different clinical applications can interact with different auto-constructed ensembles served by the same serving framework. We take a closer look at the state-of-the-art model development for the ICU application targeted in this paper, describe existing model serving platforms and optimization strategies.\\
\noindent \textbf{Deep Models for ICUs.}
Deep models, a.k.a. deep neural networks, have achieved state-of-the-art performance in many application areas due to their ability to automatically learning effective features from large scale data \cite{lecun2015deep}. ICU is one of the typical environment that generated large scale multi-modality data everyday \cite{celi2013big,bailly2018s}, so that design accurate deep models \cite{harutyunyan2017multitask} in ICU has drawn a lot of attention from researchers in recent years.

For example, in \cite{xu2018raim}, they proposed a unified recurrent attentive and intensive deep neural network for predicting decompensation and length of stay, by modeling high-frequency physiological signals, low-frequency vital signs, irregular lab measurements and discrete medication together. In \cite{wang2018supervised}, they built a recurrent neural network and learned with supervised reinforcement learning for treatment recommendation in ICU, by modeling static variables and low-frequency time-series variables. Other work applies temporal convolutions on the vital signs and lab sequences to predict decompensation \cite{nguyen2017deep}. The common characteristic of the above methods is that they aim to build unified models that serve all patients using the same data modalities. They are inflexible when encountering data heterogeneity. Besides, none of them move one step forward to serving in the real world. \\
\noindent \textbf{Deep Models Serving.}
Training is only a fraction of the end-to-end Machine Learning lifecycle and has dominated the focus of much of the literature in model development for healthcare. While great results have been achieved in terms of accuracy, not as much work focused on the difficulties and importance of efficiently deploying trained healthcare models on a platform that provides unified support for a variety of clinical use cases. 

Google's TensorFlow Serving~\cite{olston2017tensorflow} is a well-recognized open source distributed framework for Machine Learning inference and is widely used to serve deep learning models. Despite its popularity and impact, the framework has some limitations for the purposes of our target application domain. First, TensorFlow Serving provides support for single model serving. While it is possible to write additional code and integrate inference to several models, it becomes a single unit of deployment and is configured as a black box. The clinical use cases that deal with multi-modal heterogeneous models depend on first-class support for multiple models composed in a pipeline at the framework level. This is one of the key contributions of \mname. Second, models trained in other frameworks, such as PyTorch~\cite{paszke2019pytorch} cannot easily be supported by TensorFlow Serving.

Ray~\cite{moritz2018ray} is a distributed framework that provides low-level support for a combination of stateful and stateless (side-effect free) computation, suitable by design for a variety of computational patterns that appear in ML workflows, including training and serving. We build on Ray as the underlying platform and provide ML pipeline serving functionality on top---a feature Ray does not provide out of the box. Ray's tasteful choice of primitives and focus on ML and low system overhead is better suited for soft real-time serving than Apache Spark MLlib~\cite{meng2015mllib}.\\
\noindent \textbf{Optimization Strategies.}
Finally, neural architecture search \cite{zoph2016neural,liu2018progressive,cai2018proxylessnas,elsken2018neural} is an emerging area of research that targets efficient model composition at the operator level. While our work composes and serves optimal combinations of models as the fundamental building blocks. Besides, Bayesian optimization \cite{snoek2012practical,shahriari2015taking} has been used to reduce experimental costs in hyper-parameter optimization to select the best single model \cite{bergstra2011algorithms,bergstra2013making}, or building a fixed-size ensemble learning \cite{levesque2016bayesian}. It is a sequential design strategy for global optimization of black-box functions that doesn't require derivatives \cite{mockus2012bayesian}. In this paper, we solve model ensemble problem by extending Bayesian Optimization to Sequential Model-Based (Bayesian) Optimization (SMBO) \cite{hutter2011sequential} to further reduce the trial costs. Besides, we also introduce genetic algorithm to boost exploration.

\section{\mname: Health OnLine Model Ensemble Serving}
\label{sec:archdesign}
\subsection{Overview}
\label{sec:archdesign:overview}
\begin{figure}[t]
\includegraphics[width=\linewidth]{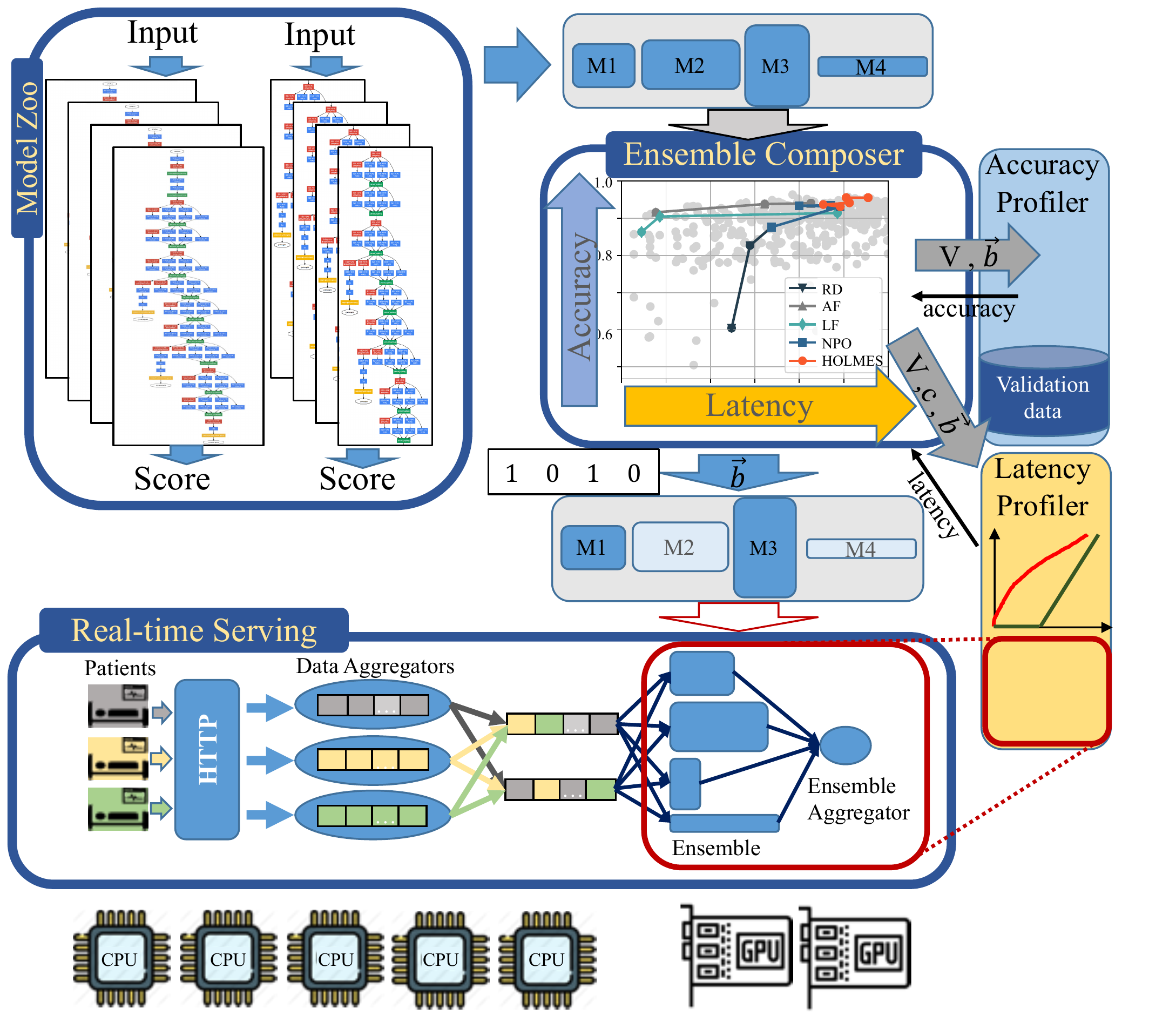}
\caption{
\mname system architecture: model zoo, ensemble composer and real-time serving system: the ensemble composer produces a model ensemble captured by $\vec{b}$; the real-time serving serves the ensemble as a mix of stateful and stateless actors connected into a queueing pipeline.}
\vspace{-1em}
\label{fig:overview}
\end{figure}

\mname consists of three main components illustrated in \Cref{fig:overview}: the model zoo, the ensemble composer, and the real-time serving system. The model zoo is populated with models trained with different model hyperparameters and different input data modalities. 

The ensemble composer optimizes for validation accuracy subject to latency constraints. It selects an optimal model set given the number of patients and the resource constraints. Then the serving pipeline is deployed with the chosen ensemble and the sensory data aggregators and is configured to operate on a large amount of streaming data and many ensemble queries in real-time.
\subsection{Model Zoo}
Model zoo is a repository of various prediction models that are already trained and ready for deployment.

ICU monitoring data are usually continuous and multimodal. They include but are not limited to dense signals like Electrocardiogram (ECG), sampled at frequencies ranging from $125$ to $1,000$ Hz, vitals signs like Blood Pressure sampled per second, and sequential discrete events like laboratory test or medication administration charted irregularly. 
Therefore, the deep learning models in model zoo are trained for each of the data modalities with short segmentation windows. A final prediction score will be generated by ensembling (i.e., aggregating predictions from multiple) individual scores across different data modalities as well as across different time segmentations within the observation window. 

The model zoo approach makes it possible to specialize in the models to different data modalities. Adding new data modalities is then simplified. Instead of retraining the whole monolithic model from scratch and replacing the model running in production completely, a new model can be added to the model zoo and \textit{automatically} chosen by the ensemble composer. The ensemble composer is triggered when any of the input conditions are changed: (a) changes in the model zoo, (b) the number of patients to serve, (c) resource constraints. 

In addition, we also consider different sizes of deep models by varying the architecture hyperparameters. Instead of training one accurate but large-sized deep model, we train a set of less accurate but smaller sized models by borrowing the prediction strength from ensemble modeling \cite{zhou2012ensemble}. By breaking models down to smaller sizes, our model zoo is more flexible and enables the feasibility that scores can serve still (even less accurate) given limited computational resources, while more accurate scores from ensemble models will be served given adequate resources. 

Models in the model zoo are trained offline using previously collected ICU data, tuned and validated on an independent data from new patients. We store the pre-trained models along with their profiles. As shown in Table \ref{tb:model_profile}, one example model profile $\bm{v} \in \mathbb{R}^m$ ($m$ as the number of fields specified in a profile) can contain model size details such as depth, width, number of multiply and accumulate operations and GPU memory usages, input data information such as data modality and length of data segmentation that models were trained on, and model performance on the validation set such as ROC-AUC scores, etc. Fields in model profiles can vary case by case. Table \ref{tb:model_profile} in the Appendix lists those used in this paper.

\begin{table}[t]
\centering
\caption{Notations. }
\resizebox{0.45\textwidth}{!}{
\begin{tabular}{l|l}
\toprule
\textbf{Notation} & \textbf{Definition}  \\
\midrule
$n$ & Number of models in model zoo \\ 
$m$ & Number of fields for model description \\ 
$d$ & Number of fields for system configuration \\ 
$\mathcal{M}=\{m_1, m_2, ..., m_n\}$ & Model zoo \\
$\bm{V} \in \mathbb{R}^{n \times m}$ & Model description  \\
$\bm{c} \in \mathbb{R}^{d}$ & System configuration \\
$\mathcal{B} \triangleq \{0,1\}^n$ & Exploration space \\
$\bm{B}$ & Profile set \\
$\bm{b} \in \mathcal{B}$ & Model selector \\
$f_a(\bm{V}, \bm{b})$ & Accuracy profiler \\
$f_l(\bm{V}, \bm{c}, \bm{b})$ & Latency profiler \\
\bottomrule
\end{tabular}
}
\label{tb:notations}
\end{table}

\subsection{Ensemble Composer}
\label{sec:archdesign:algos}
In this section we describe the second component of \mname{}---the ensemble composer.
Model composition into ensembles has been known to improve performance, but focused primarily on accuracy~\cite{zhou2012ensemble}. Unconstrained, ensemble size can easily absorb all models in the model zoo (Sec.~\ref{sec:archdesign:overview}). The latency of deploying large ensembles on limited computational resources can be prohibitive, particularly for real-time serving scenarios.

\begin{figure}[ht]
\includegraphics[width=\linewidth]{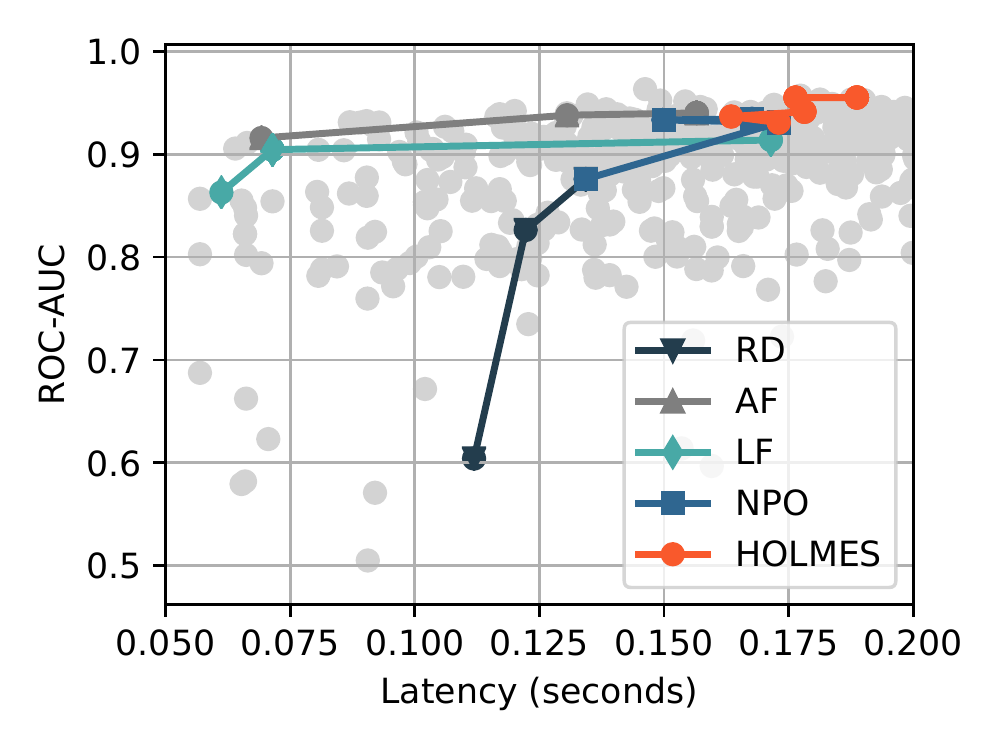}
\caption{ROC-AUC vs. Latency vary by iterations in different Explore algorithms}
\label{fig:explore_app}
\end{figure}

To address this, we propose a latency-aware ensemble composition method.
In Fig.~\ref{fig:explore_app}, we show that the range of possible latency/accuracy outcomes for arbitrary ensembles drawn from the model zoo is large. The best performing method must achieve the highest accuracy for \textbf{any} specified latency threshold. 
Below we formulate ensemble composition as an optimization problem and propose an efficient search algorithm within Bayesian optimization framework to find an optimal solution.

\subsubsection{Problem setup}

Suppose we have $M_1$, $M_2$,..., up to $M_n$ models in the model zoo with each model having a profile $\bm{v}_i \in \mathbb{R}^m$, then we can represent the entire model zoo as $\bm{V} \in \mathbb{R}^{n \times m}$. We denote one system configuration as a vector $\bm{c} \in \mathbb{R}^{d}$, where $d$ is the total number of items needed to be configured in a system.
They include but are not limited to the number of GPUs, memory size, or number of clients in the system. 
Lastly a model ensemble is uniquely identified
using a binary indicator vector $\bm{b} \in \mathcal{B} \triangleq \{0,1\}^n$ such that $b_i = 1$ indicates model $M_i$ is selected for ensembling and $0$ otherwise. 

\noindent \textbf{Accuracy and latency trade-off.}
Our goal of ensemble composer is to find the $\bm{b}^*$ such that
\begin{equation}
\begin{aligned}
\bm{b}^* \equiv \argmax_{\bm{b} \in \{0,1\}^{n}} \quad & \underbrace{f_a(\bm{V}, \bm{b})}_{\text{Accuracy profiler}} \\
\textrm{s.t.} \quad & \underbrace{f_l(\bm{V}, \bm{c}, \bm{b})}_{\text{Latency profiler }} \leq L, \\
\end{aligned}\label{eq:latency}
\end{equation}
where $f_a(\bm{V}, \bm{b})$ is an \textit{accuracy profiler} that produces prediction accuracy on the validation set given an model ensemble $\bm{b}$ selected from the model zoo $\bm{V}$, $f_l(\bm{V}, \bm{c}, \bm{b})$ is a \textit{latency profiler} that computes the latency of serving the ensemble $\bm{b}$ under a certain system configuration $\bm{c}$, and $L$ is the latency constraint required by a real-time serving system. We introduce a function $\delta$ and rewrite the above optimization problem in the following equation 
\begin{equation}
\begin{aligned}
\max_{\bm{b} \in \{0,1\}^{n}} \quad & L_a(\bm{b}) = f_a(\bm{V}, \bm{b}) + \delta(L-f_l(\bm{V}, \bm{c}, \bm{b})),\\
\end{aligned}\label{eq:opt1}
\end{equation}
where $\delta$ is an activation function. If $\delta$ is a linear function, then the above Eq. (\ref{eq:opt1}) converges to the common Lagrange multiplier setting that supports soft constraint. If $\delta$ is a step function specified below, we reach a hard constraint on latency instead: 
\begin{equation}
\begin{aligned}
\delta(x) =
\begin{cases}
-\inf, & if \enskip x < 0 \\
0, & otherwise
\end{cases}\\
\end{aligned}
\end{equation}

Alternatively, we can switch the objective function and constraint in Eq. (\ref{eq:latency}) and minimize the latency subject to a minimum accuracy requirement given any accuracy sensitive task. It's beyond the scope of this paper, but its formulation is similar (listed in the supplementary material) and can be solved using the same searching algorithm we introduce below.

\subsubsection{Ensemble Composer exploration}
There exist two main challenges for solving the Eq. (\ref{eq:opt1}) optimization problem: 1) unknown accuracy profiler $f_a(\bm{V}, \bm{b})$ and latency profiler $f_l(\bm{V}, \bm{c}, \bm{b})$ for all ensemble choices when model zoo size $n$ is large ($|\bm{b}| = 2^n$!); and 2) high dimensional binary searching space as opposed to continuous. 

To tackle the first challenge, we adapt the widely known black-box searching algorithm -- Bayesian optimization -- into our searching algorithm, in which surrogate probability models \cite{koziel2013surrogate} are used for approximating the accuracy and latency profilers. For the second challenge, as typical Bayesian exploration functions (such as Gaussian process regression) usually searches parameters in continuous space, we propose to use Genetic search algorithm \cite{whitley1994genetic} for exploration so that our method can support efficient search in high-dimensional binary space.

Under the Bayesian Optimization framework, our goal is to iteratively enrich a true valued set $\bm{B} \in \mathcal{B}$, in which accuracy and latency are truly profiled by $f_a(\bm{V}, \bm{b})$ and $f_l(\bm{V}, \bm{b})$, and then update the objective $L_a(\bm{b})$ in Eq. (\ref{eq:opt1}) given all the available $\bm{b}$'s in the set. When it reaches the budget of $N$ profiler calls, a $\hat{\bm{b}}^*$ that maximizes $L_a(\bm{b})$ over the current set $\bm{B}$ is returned as the optimal solution. The crucial step in the framework is how to smartly explore the searching space at each iteration, and pick the right vector into set $\bm{B}$ so that we don't waste the budget of calling the profilers for evaluating the true accuracy and latency. \\
\noindent \textbf{a) Genetic algorithm for exploring binary parameter space.}
The key idea of exploration is to search new $\bm{b}$'s towards the optimal $\bm{b}^*$ and efficiently enlarge the set $\bm{B}$. Usually, when searching domain is continuous \cite{bergstra2012random}, random exploration would be better than grid search. However, when searching domain is binary as in our case, the benefit of randomness from dimensionality combination is low -- there are only two values for each dimension of $\bm{b}$. Thus, we borrow the ideas from Genetic Algorithm (GA) \cite{whitley1994genetic}, in which each $\bm{b}$ in the explore space $\mathcal{B}$ can be naturally represented as a genotype. Hence, genetic operators such as recombination and mutation can be utilized to explore our binary parameters. In detail, we define
\begin{equation}
\begin{aligned}
\text{Recombination($\bm{b}_1$, $\bm{b}_2$):} & \quad \bm{b} \triangleq concate(\bm{b}_1[:i], \bm{b}_2[i+1:]), \\
\text{Mutation($\bm{b}_3$, $1$):} & \quad \bm{b} \triangleq \bm{b}_3[i] = 
\begin{cases}
1, & if \enskip \bm{b}_3[i] = 0 \\
0, & if \enskip \bm{b}_3[i] = 1,
\end{cases}
\end{aligned}
\end{equation}
where $i$ is a random number from $\{1,2,\dots,n\}$. We perform $S$ times of mutations (called mutation degree)  on $\bm{b}_3$, i.e., randomly sample a new vector from the neighborhood of $\bm{b}_3$ within a Manhattan distance of $S$, and denote the newly formed candidate set as $\bm{B}'$.

\noindent \textbf{b) Surrogate models for approximating accuracy and latency profilers.}
For selecting the right points into $\bm{B}$, we aim to build less expensive surrogate probability models for approximating $f_a(\bm{V}, \bm{b})$ and $f_l(\bm{V}, \bm{c}, \bm{b})$. That is, based on the current true value set $\bm{B}$, we fit separated latency surrogate probability model $\hat{f}_l$ using $\{(\bm{b}, f_l(\bm{V}, \bm{c}, \bm{b})):$ for all $\bm{b} \in \bm{B}\}$, and a accuracy surrogate probability model $\hat{f}_a$ using $\{(\bm{b}, f_a(\bm{V}, \bm{b})):$ for all $\bm{b} \in \bm{B}\}$.
Given a new candidate $\bm{b}' \in \bm{B}'$, we can evaluate their approximated latency as $\hat{f}_l(\bm{V}, \bm{b}')$ and approximated accuracy as $\hat{f}_l(\bm{V}, \bm{b}')$ using the surrogate models. Then we pick $K$ candidates whose approximated objective values computed by Eq. (\ref{eq:opt1}) have ranked top K among the set $\bm{B}'$, add them into the current set $\bm{B}$ and evaluate their true values using the accuracy and latency profilers.

We summarize our algorithm in Algorithm \ref{alg:proxy} (pseudo code of function \textit{Explore} is given in supplementary material Algorithm \ref{alg:explore}).
\begin{algorithm}[t]
\caption{Ensemble Composer exploration in \mname }
\label{alg:proxy}
\begin{algorithmic}[1]
\STATE \textbf{Input}: $\bm{V}$  model zoo, $\bm{c}$ system constraint, $L$ latency constraint
\STATE \textbf{Parameters}: $\lambda$, $\bm{N}$ number of search iterations, $\bm{N_0}$ number of warm start samples, $\bm{M}$ number of explore samples, $\bm{K}$ number of newly added samples for profiling, $\bm{S}$ degree of mutation, $\bm{p}$ probability of genetic explore, $\bm{q}$ probability of mutation. 
\STATE \textbf{Output}: $\bm{b}^*$
\STATE Initialize surrogate models $\hat{f}_a$ and $\hat{f}_l$
\STATE \textcolor{blue}{\texttt{/* Warm start to get some seed solutions */}}
\STATE Warm start to get an initial $\bm{\bar{B}} \in \{0,1\}^{N \times |V|}$
\STATE $\bm{B}=\emptyset \bigcup \bm{\bar{B}}$, $Y_a=\emptyset$, $Y_l=\emptyset$
\FOR {i=1:$N$}
\STATE \textcolor{blue}{\texttt{/* Profile accuracy and latency in $\bm{B}$ */}}
\STATE Profile $\bm{\bar{Y}}_a= \{ f_a(\bm{V}, \bm{b}), \forall \bm{b} \in \bm{B}\}$, $\bm{\bar{Y}}_l=\{f_l(\bm{V}, \bm{c}, \bm{b}), \forall \bm{b} \in \bm{B}\}$
\STATE $\bm{Y}_a=\bm{Y}_a \bigcup \bm{\bar{Y}}_a$, $\bm{Y}_l=\bm{Y}_l \bigcup \bm{\bar{Y}}_l$
\STATE \textcolor{blue}{\texttt{/* Fit surrogate models on profiled results */}}
\STATE Fit $\hat{f}_a$ using $\bm{B}$ and $\bm{Y}_a$, and fit $\hat{f}_l$ using $\bm{B}$ and $\bm{Y}_l$
\STATE \textcolor{blue}{\texttt{/* Genetic exploration, details in Algo.\ref{alg:explore} */}}
\STATE $\bm{B}'=Explore(\bm{B}, M, S, p, q)$
\STATE \textcolor{blue}{\texttt{/* Approx. accuracy and latency in $\bm{B}'$  */}}
\STATE Approx. $\hat{L}_a(\bm{B}')  = \{ \hat{f}_a(\bm{V}, \bm{b}) + \lambda(L-\hat{f}_l(\bm{V}, \bm{c}, \bm{b})), \forall \bm{b} \in \bm{B}' \}$
\STATE \textcolor{blue}{\texttt{/* Pick top-K highest valued vectors to get $\bm{\bar{B}}$ */}}
\STATE $\bm{\bar{B}}=\text{argsort}_K(\hat{L}_a(\bm{B}'))$
\STATE \textcolor{blue}{\texttt{/* Add seed solutions to profile set */}}
\STATE $\bm{B}=\bm{B} \bigcup \bm{\bar{B}}$
\ENDFOR
\STATE \textcolor{blue}{\texttt{/* Get the optimal solution from $\bm{B}$ */}}
\STATE $\bm{b}^* = \argmax_{\bm{b} \in \bm{B}} L_a(\bm{b})$
\end{algorithmic}
\end{algorithm}
\subsubsection{Prediction ensemble}
Given the optimal solution $\bm{b}^*$, we make our final prediction using  bagging ensemble \cite{bagging}:
\begin{equation}
\mathbb{E}[Y | \bm{x}] = \frac{1}{n}\sum_{i=1}^n \bm{b}^*_i \mathbb{E}_{m_i}[Y | \bm{x}],
\end{equation}
where $Y$ is the outcome measure of interest for a given prediction task, and $\bm{x}$ is an instance of ICU data input.
\subsection{Real-time Serving}
\label{sec:archdesign:serve}
\begin{figure}
\includegraphics[width=\linewidth]{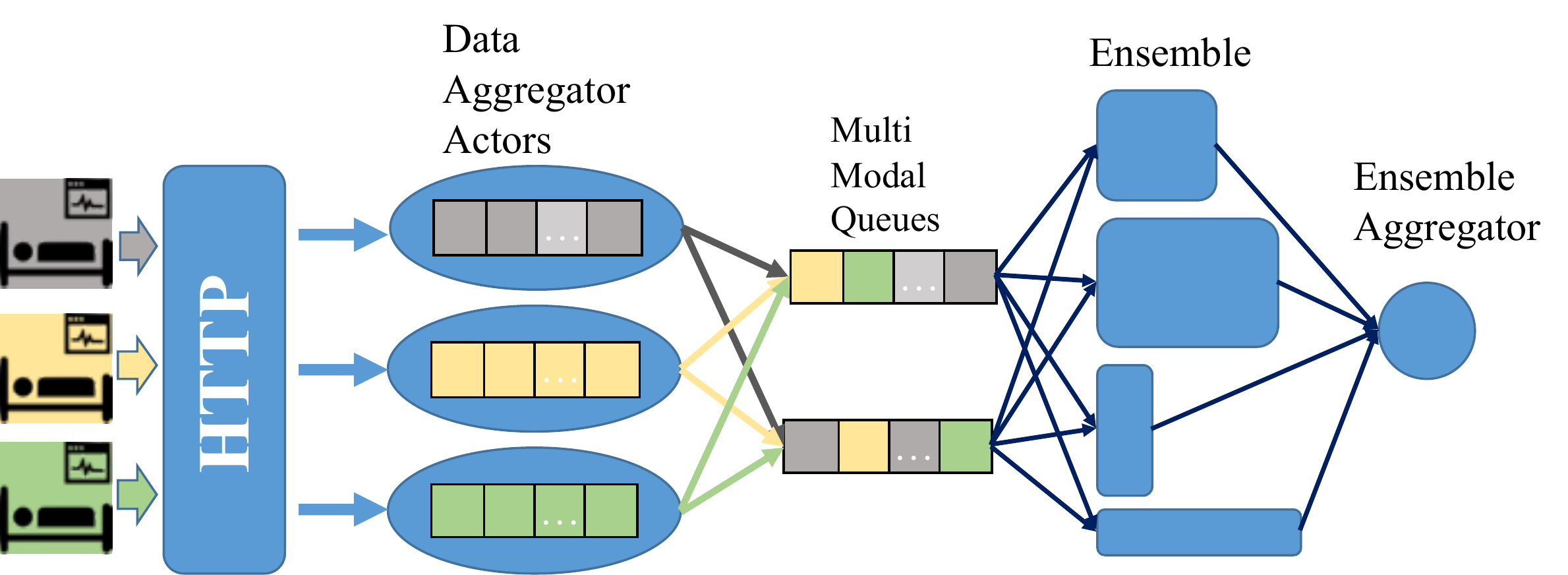}
\caption{\mname Ensemble Serving System: requires a combination of stateful and stateless components, buffers multi-modal and multi-frequency sensory data in data aggregators, conditionally queries the ensemble, and returns the aggregated ensemble result to the patient monitoring system.}
\label{fig:archidesign:serve}
\vspace{-1.em}
\end{figure}
{\bf Overview.}
\label{sec:archdesign:serve:overview}
The third component of \mname is the real-time serving system used to serve the model ensemble as part of the end-to-end ensemble pipeline. It serves two purposes: (1) ensemble candidate latency profiling (Fig.~\ref{fig:overview}) and (2) real-time ensemble pipeline serving.

The served pipeline consists of four main components: the source of streaming data (typically bed-side patient monitoring data), the HTTP server that simplifies data ingest into the serving system, patient data accumulators that buffer the data, and the ensemble itself. 
The \mname pipeline is implemented by deploying the data aggregators and ensemble models as actors on top of Ray~\cite{moritz2018ray}.
The ensemble queries are routed through multi-modal queues, with each queue corresponding to the appropriate data modality.

{\bf Support for stateful compute.} The aggregator actors make it possible to deliver synchronized, coordinated buffers of streaming data to the ensemble.
While the  data  stream may come at different frequencies and modalities, the input supplied to the ensemble must correspond to the same observation window across all sensors (to capture sensory correlations). E.g., ECG produces waveforms at 250 samples per second (250qps), while BP monitor outputs 1 sample per second (1qps). The data aggregator actors buffer this data for the \textit{same} interval of time $\Delta{T}$ before the ensemble is queried. This requires a platform that's capable of supporting stateful compute.

{\bf Support for stateless compute.} Ensemble models are deployed as actors with a queue. 
Inference performed on the ensemble models implies that the models themselves can be stateless. This simplifies model lifecycle management and deployment and makes it possible to horizontally scale individual ensemble models as the load on the system increases. It is, therefore, imperative that the serving platform natively supports a combination of both \textit{stateful} and \textit{stateless} compute---a property \mname borrows from Ray~\cite{moritz2018ray} in its implementation of the ensemble serving platform.

{\bf Latency profiling.}
The latency profiling component is exposed to the ensemble composer (Sec.~\ref{sec:archdesign:algos})
through the API $f_l(V, c, \vec{b_i})$, where $\vec{b_i}$ corresponds to the model zoo subset choice considered at iteration $i$. Given the frequency of interaction with the latency profiler, it must be highly performing. 
Ensemble serving latency reported by the latency profiler consists of two constituent components: the queueing delay $T_q$ and the serving latency of the ensemble $T_s$. Thus, 
$\hat{T} = T_q + T_s $,
where $\hat{T}$ is the estimate of end-to-end response time,
$T_q$ denotes queueing delays in the system, and
$T_s$ denotes serving delay of the ensemble.
This breakdown is fundamental as $T_q$ depends on the characteristics of the client load (i.e., ingest rate, number of clients, inter-arrival process characteristics), while $T_s$ depends on the characteristics of the ensemble and the resources used. As such, the methodology for estimating $T_q$ and $T_s$ is completely different.

{\bf Estimating $T_s$ and $T_q$.} 
To estimate ensemble latency $T_s$, we measure its throughput capacity $\mu$ (qps) by directly performing inference on the ensemble in a closed-loop fashion and averaging over a statistically significant number of queries. $T_s$ is then estimated by configuring the clients to generate queries at the ingest rate $\lambda \leq \mu$ and taking the $95^{th}$ \%-ile latency. This provides enough information to estimate $T_q$.

\begin{figure}[!ht]
\includegraphics[width=0.45\linewidth]{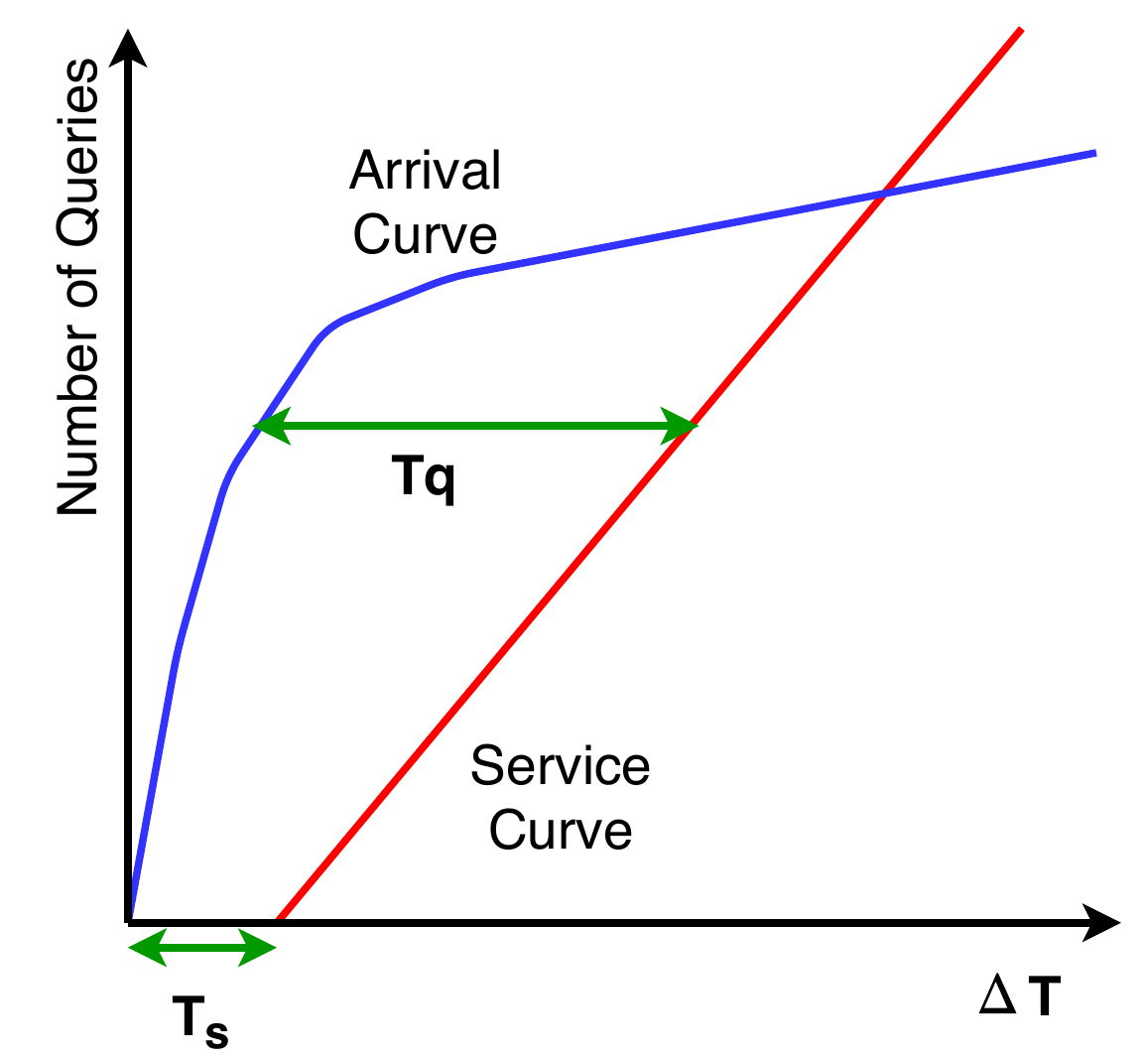}
\vspace{-1em}
\caption{\mname latency estimator: latency estimation leverages network calculus. The arrival curve is constructed from query data observed at runtime during profiling. The service curve is constructed analytically. The maximum distance between the curves is known as a tight upper bound on queueing latency $T_q$.}
\label{fig:latency:estimate}
\vspace{-1.5em}
\end{figure}

$T_q$ depends on query inter-arrival distribution and, thus, indirectly depends on the number of patients generating query load on the ensemble. To estimate $T_q$ we rely on network calculus. As profiling queries are generated, we construct two curves: an arrival and service curve. 
The arrival curve captures the max number of queries that have thus far been observed within any time interval of length $\Delta{t}$. The service curve captures the number of queries that can be serviced within an interval of time $\Delta{t}$-long. The maximum horizontal distance between these two curves (Fig.~\ref{fig:latency:estimate}) is a known tight upper bound on the queueing latency for such a system. 

\section{Experimental evaluation }
\label{sec:exp}
\subsection{Experimental Setup}
\label{sec:exp:setup}
\subsubsection{Model zoo training}
\label{sec:exp:setup:training}
\label{sec:exp_zoo}
Given a specific prediction task, model zoo needs to be populated with pre-trained models for subsequent ensemble composition. Here we describe what prediction task is targeted in our experiments, what ICU data is used, and what/how candidate models are trained.

\noindent \textbf{Task. } We focus on a binary classification task that predicts if a post-surgical (specifically the Norwood surgery) patient is getting \textit{stable} or stay still \textit{critical} in the Cardiac Intensive Care Units (CICUs).

\noindent \textbf{Data. }
We have collected multimodal ICU data from 57 children who had undergone the Norwood procedure between 2016/10 and 2019/09 in the Cardiac Intensive Care Unit (CICU) at Children Healthcare of Atlanta (CHOA). These patients stayed at the CICU after the operation from 4 days up to 30 days, with 45 (78.9\%) being successfully discharged to the general care cardiology floor, 6 (10.5\%) deceased and 6 (10.5\%) transferred for other operations. We extract data from the first two days of post-op CICU stays from all the 57 patients, and label them as 0 (\textit{critical}); we extract data from the last day prior to floor transfer from the 45 successfully discharged patients, and label them as 1 (\textit{stable}). 

Modalities in this data include 3-lead (I, II, and III) ECG waveforms sampled at 250 Hz, 7 vital signs (mean blood pressure, SpO2, etc) sampled per second and 8 discrete lab results (pH, lactic acid, etc) measured when needed. We segment the continuous signals (waveforms and vital signs) into 30 second clips, and result at vectors of length 7,500 per single lead waveform and length 30 per single vital sign. There are in total of $328,320$ data points per each signal labelled as 0 and $129,600$ per signal labelled as 1.  

\noindent \textbf{Training details. } We first split the cohort by putting 47 earlier patients into training set and 10 recent patients into test set. We train a state-of-art convolutional neural network, by modifying the kernel in the convolutional layer in ResNeXt \cite{xie2017aggregated} from 2-D patch to 1D stripe, individually for each single lead ECG clips. By filtering out missing signals, we obtain 164,972 training samples and 71,342 validation samples for lead-I ECG, 230,046 training and 71,364 validation samples for lead-II, and 130,564 training and 60,785 validation samples for lead-III. 

We train differently sized networks per each ECG lead, particularly varying the number of filters in the first convolutional layer of the network between $\{8, 16, 32, 64, 128\}$ and number of residual blocks between $\{2, 4, 8, 16\}$. As a result, we reach at a deep model zoo of size $|\bm{V}| = 3 \text{ (ECG leads)} \times 5 \text{ (filters) } \times 4 \text{ (blocks) } = 60$. 
As there is no need to train deep models on low resolution data like vital signs or lab measurements, we simply train a random forest for each vital sign, and a Logistic regression for labs. 
Since inference from these models using CPUs is negligible compared to inference from deep models using GPUs, we do not include them into the model zoo and don't take their inference time into account for system latency. But prediction accuracy ensembles the optimal deep models selected from the model zoo with these ML models.

\subsubsection{System setup details}
We specify system configuration $\bm{c} \in \mathbb{R}^{2}$ to include number of GPUs and number of patients. We use 2 NVIDIA Tesla V100s with 32GB RAM as ensemble serving node in our experimental setup. In order to simulate the client request, we build a data generator to simulate the data flow in ICU  which generate requests at a frequency of 250qps from a client node. We connect the serving node and client node via HTTP and RPC. 
The serving node can start RPC call to client node to start a simulation to generate client requests. The data generated will then be sent by the client node and captured by the HTTP server. The data captured is passed on and stored inside accumulators running on the serving node (Figure \ref{fig:overview}). When the observation window is reached the data will then be further passed on for inference to the ensemble queue. 
The ensemble models dequeue queries and perform prediction on query data.
The end-to-end latency that we profile is the latency of serving system from the moment the data is captured by HTTP server until the prediction results are finished by the ensemble \footnote{Our code can be found at \url{https://github.com/hsd1503/HOLMES} }.

\subsection{End-to-end \mname Performance}
\label{sec:exp:algos}
In this section we evaluate the performance of \mname' ensemble composition component with respect to the prediction accuracy and latency.
We report prediction accuracy in terms of area under receiver operating characteristic curve (ROC-AUC), area under precision recall curve (PR-AUC), F1 score, and accuracy. We report system efficiency in terms of latency (second). We compare \mname with the following baselines:
\begin{itemize}[leftmargin=5mm]
\itemsep0em
\item \textbf{Random (RD)} iteratively chooses one random single model from model zoo without replacement, and adds it to the current model set, till the ensemble model exceeds latency constraint. 
\item \textbf{Accuracy First (AF)} iteratively chooses the next most accurate single model from model zoo, and adds it to the current model set, till the ensemble model exceeds latency constraint. 
\item \textbf{Latency First (LF)} iteratively chooses the next lowest latency single model from model zoo, and adds it to the current model set, till the ensemble model exceeds latency constraint. 
\item \textbf{Non-Parametric Optimization (NPO)} (modified based on  \cite{snoek2012practical}) iteratively chooses a random subset $\bm{B}$ (size bounded by the number of models selected by LF) from model zoo, and merges them to the current model set, till the number of profiler calls exceeds the budget $N$; returns the $\bm{b}^*$ that maximizes the objective function in Eq. (\ref{eq:opt1}) over the final explored model set.
\end{itemize}

For both NPO and \mname, we also add solutions from RD, AF and LF as their initial profiling. The budget $N$ to profiler calls is the same for NPO and \mname. In \mname,  we build two random forest \cite{breiman2001random} as the surrogate models for accuracy and latency. 

\subsubsection{Overall performance compared with baselines}
Table \ref{tb:compare} summarizes the prediction performance from all methods under a 200ms latency constraint.
\mname achieves the best performance for all measurements within the same latency constraint.
\begin{table}[ht]
\caption{Comparison results.   }
\vspace{-1em}
\label{tb:compare}
\resizebox{0.49\textwidth}{!}{
\begin{tabular}{lcccc}
\toprule
Method & ROC-AUC & PR-AUC & F1 & Accuracy \\
\midrule
RD & 0.8758 $\pm$ 0.1334 & 0.8198 $\pm$ 0.2404 & 0.6887 $\pm$ 0.2246 & 0.7760 $\pm$ 0.1311 \\
AF & 0.9307 $\pm$ 0.0862 & 0.9025 $\pm$ 0.0791 & 0.7426 $\pm$ 0.2920 & 0.8526 $\pm$ 0.1113 \\
LF & 0.9135 $\pm$ 0.1020 & 0.8755 $\pm$ 0.1093 & 0.8302 $\pm$ 0.1387 & 0.8695 $\pm$ 0.1083 \\
NPO & 0.9343 $\pm$ 0.0741 & 0.9078 $\pm$ 0.1418 & 0.8237 $\pm$ 0.1828 & 0.8756 $\pm$ 0.0941 \\
\textbf{\mname} & \textbf{0.9551 $\pm$ 0.0521} & \textbf{0.9349 $\pm$ 0.0834} & \textbf{0.8501 $\pm$ 0.1054} & \textbf{0.8837 $\pm$ 0.0815} \\
\bottomrule
\end{tabular}
}
\vspace{-1em}
\end{table}
\subsubsection{Tradeoff space exploration efficacy}
\begin{figure}[t]
\begin{tabular}{cc}
\hspace{-3mm}\includegraphics[width=0.5\linewidth]{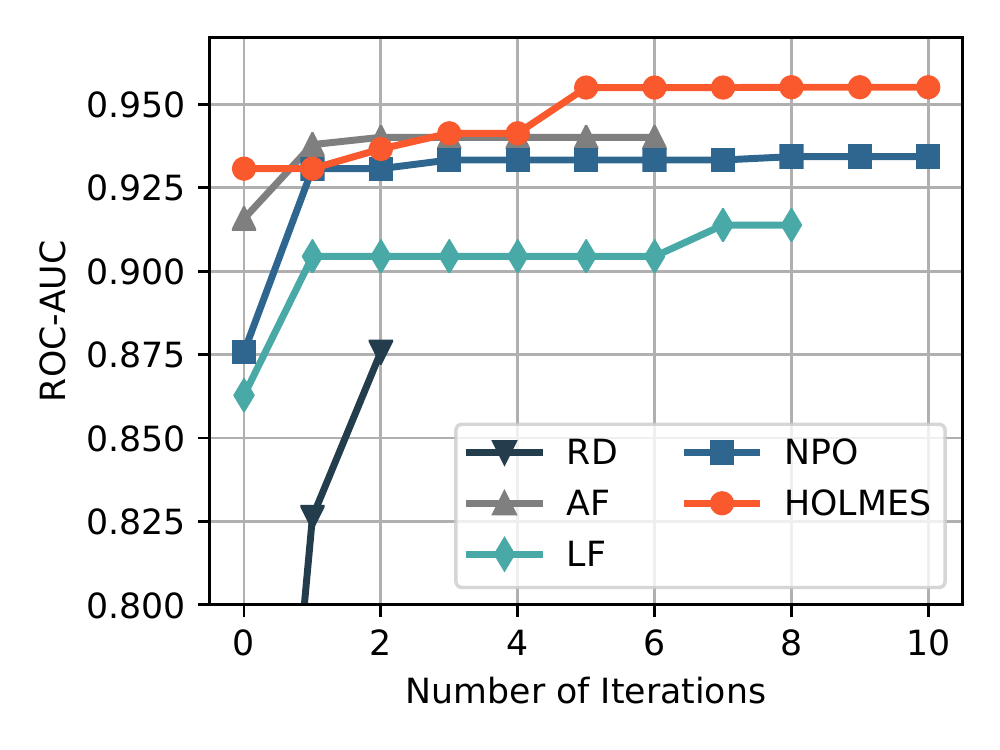} &
\hspace{-3mm}\includegraphics[width=0.5\linewidth]{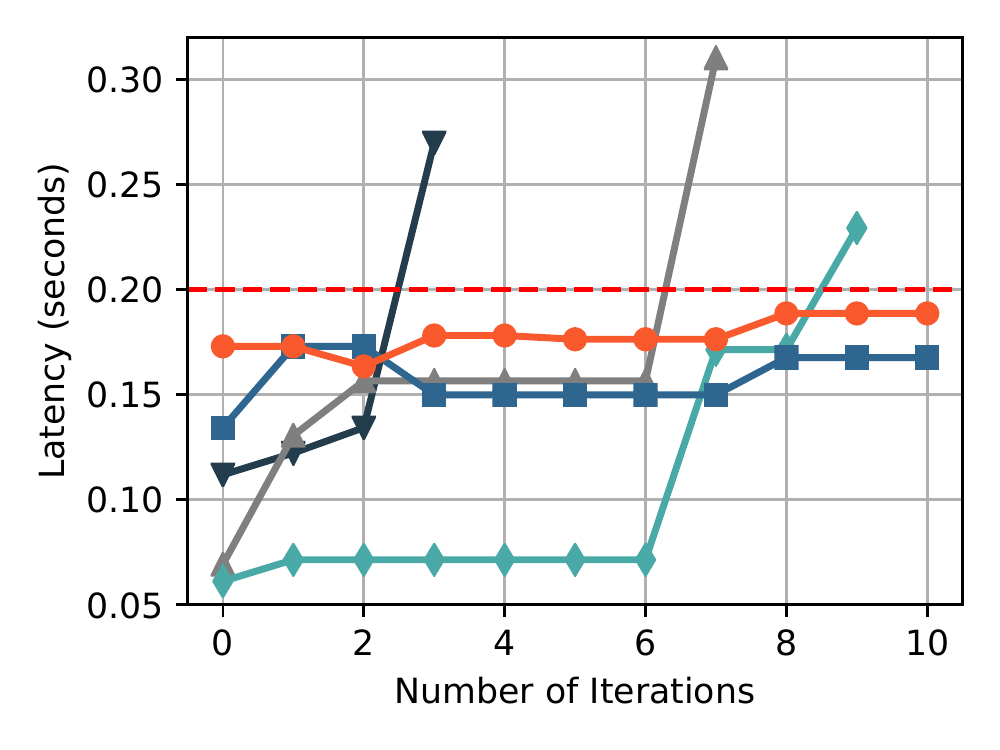} \\
\end{tabular}
\vspace{-1em}
\caption{Search trajectory: accuracy (left) and latency (right) as a function of iteration.  }
\label{fig:explore1}
\vspace{-1em}
\end{figure}

First, we track the search trajectory through the accuracy/latency tradeoff space in Figure \ref{fig:explore1}, tracking incremental changes in accuracy (Left) and latency (Right).
\mname is able to quickly reach the 200ms latency constraint, but continuing model selection, packing a more accurate model ensemble within the latency budget.
RD, AF, and LF stop after exceeding the latency budget (higher than the 200ms horizontal line) and don't reach optimality w.r.t. AUC-ROC scores. NPO search trajectory stays under the latency budget, but doesn't reach optimal ROC-AUC, due to inefficient random exploration.

\begin{figure}[t]
\begin{tabular}{ccc}
\hspace{-3mm}\includegraphics[width=0.35\linewidth]{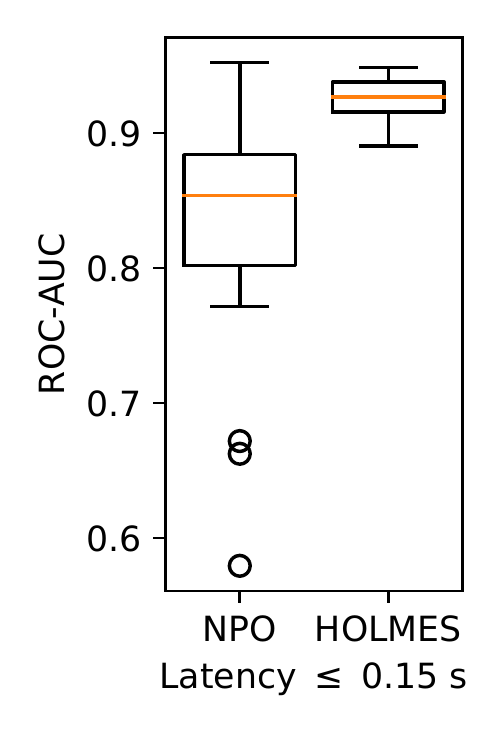} &
\hspace{-5mm}\includegraphics[width=0.35\linewidth]{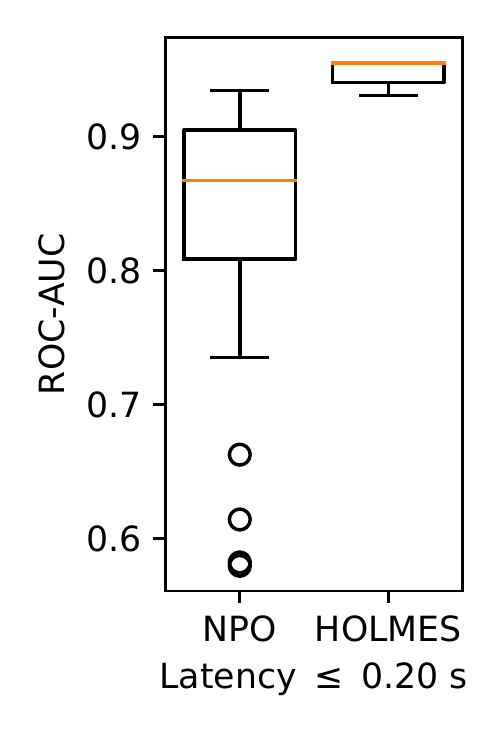} &
\hspace{-5mm}\includegraphics[width=0.35\linewidth]{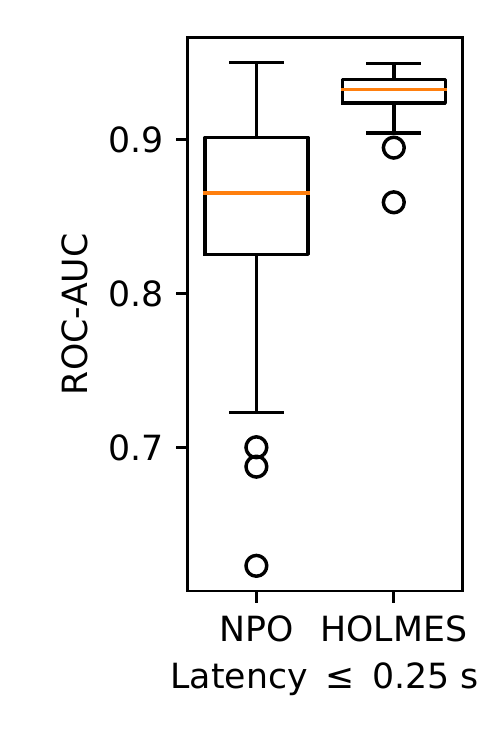} \\
\end{tabular}
\vspace{-1.5em}
\caption{ \mname overall finds more accurate ensemble models than NPO under different latency constraints.  }
\label{fig:vary_latency}
\vspace{-1.em}
\end{figure}

Second, we show that \mname finds ensembles with accuracy/latency on the Pareto frontier of the tradeoff space.
Namely, for a range of fixed latency budgets, \mname consistently composes ensembles that outperform NPO (the highest performing baseline) w.r.t. ROC-AUC (Figure \ref{fig:vary_latency}). Furthermore, NPO ROC-AUC variance is higher due to the unstable random search/exploration. In contrast \mname produces narrower ROC-AUC distribution as the ensemble models it explores have smaller ROC-AUC variance.

\begin{figure}[t]
\hspace{-3mm}\includegraphics[width=0.6\linewidth]{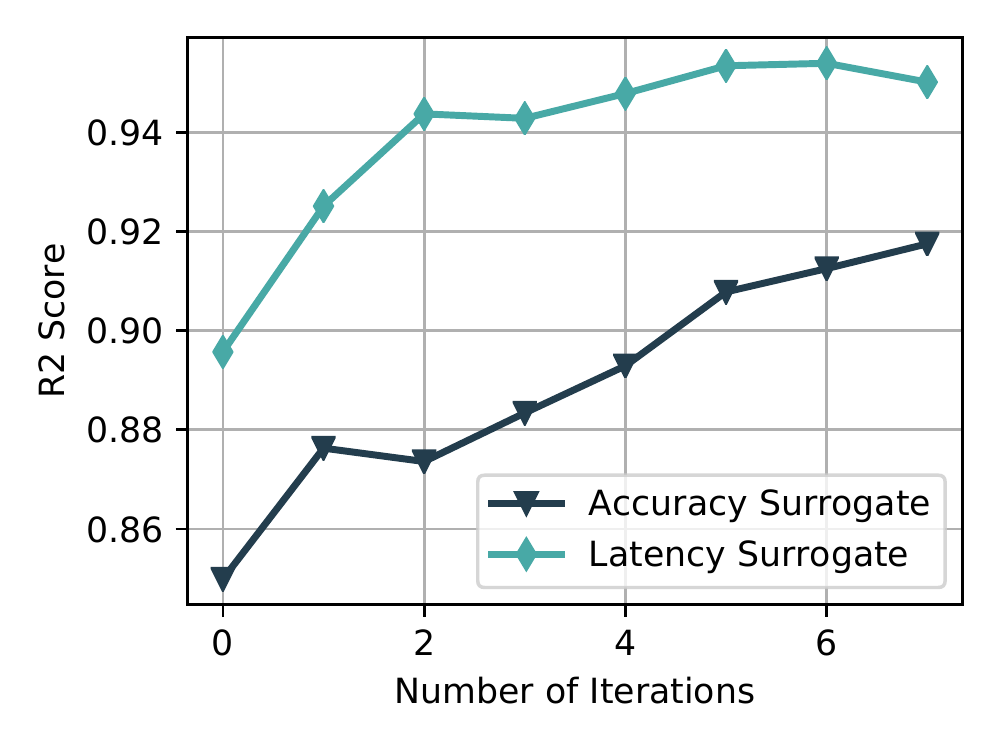}
\vspace{-1.5em}
\caption{ Surrogate models' performances increasing with the number of profiler interactions.  }
\label{fig:surrogate}
\vspace{-1em}
\end{figure}

Third, we validate the efficacy of our exploration algorithm in \mname by showing the two surrogate models improve when the number of exploration steps increases. In Figure \ref{fig:surrogate}, we plot $R_2$ score at each iteration step. $R_2$ measures the differences between the predicted accuracy/latency by the surrogate model and the true accuracy/latency profiled by the profiler on an independent validation set that has not been explored by the algorithm. We can tell both models' predictions get improved as their $R_2$ scores increase. This result explains why \mname finds more accurate ensemble models than NPO, even if they have the same number of iterations -- \mname can explore and predict more promising ensemble models using surrogates before actually profiling them. 
\subsection{Ensemble Serving System Performance}
\label{sec:exp:sys}
\subsubsection{Online vs. offline inference latency}
\label{sec:exp:sys:lat}
\begin{figure}[ht]
    \centering
    \hspace{-5mm}\includegraphics[width=\linewidth]{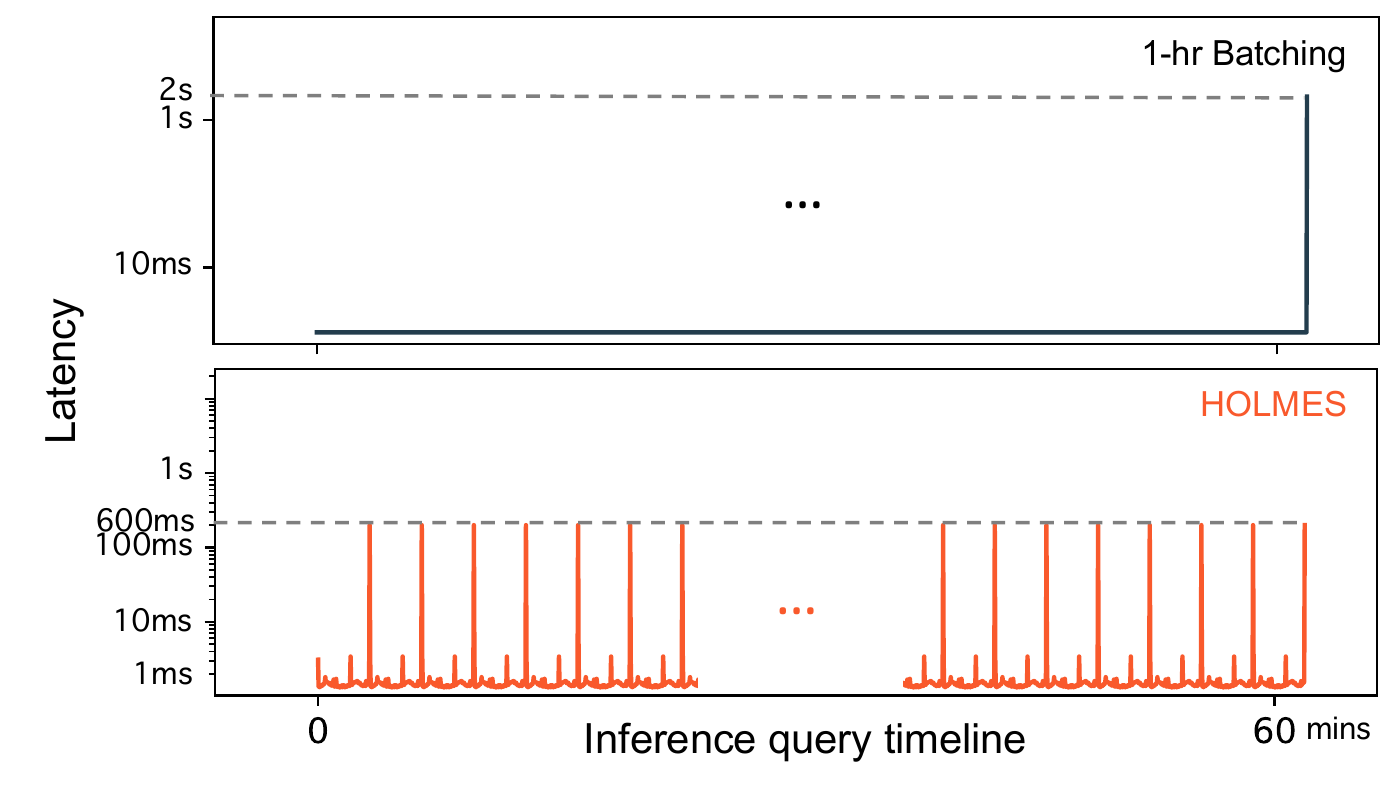}
    \vspace{-1.5em}
    \caption{Comparison of an end-to-end timeline from the batching solution (every 1 hour) and \mname.}
    \label{fig:expected_trace_e2e}
    \vspace{-1em}
\end{figure}

We now evaluate the serving system performance.
First, we demonstrate the benefit of using an online serving system relative to the conventionally adopted approach of periodically re-evaluating patient condition offline. In Figure \ref{fig:expected_trace_e2e}, we plot the ensemble query latency over time for \mname and the conventional batching method commonly adopted in hospital ICUs for patient discharge evaluation.

The experiment is carried out for a single patient over the period of 60 minutes. There's a single periodic spike at the end of the 60min time window corresponding to the latency of evaluating patient's condition on accumulated stale data. This approach affects both the accuracy (see Figure \ref{fig:intro2}) and latency. It can be seen that the batching approach incurs inference latency that's an order of magnitude higher than \mname (log y-axis).

\mname, on the other hand performs evaluation every 30s, corresponding to the periodic spikes up to an order of magnitude of $10^{-1} s$. The smaller latency in between the spikes correspond to just the sensory data collection part of the pipeline (Figure~\ref{fig:archidesign:serve}). As the system is well-provisioned, we expect the online sensory data collection and aggregation latency not to diverge and remain approximately at the same level, in this case, at the order of magnitude of 1-10ms.
For the purposes of this evaluation, the highest accuracy model was chosen as the prediction model.

Thus, we observe that by reducing data staleness, we not only gain in terms of accuracy but also achieve
an order of magnitude faster inference latency for patient condition re-evaluation.
\subsubsection{Varying resource constraints}
\label{sec:exp:sys:res}
We use the ensemble models selected by \mname to perform the scalability experiments. 

\begin{figure}[ht]
\centering
    \includegraphics[width=0.8\linewidth]{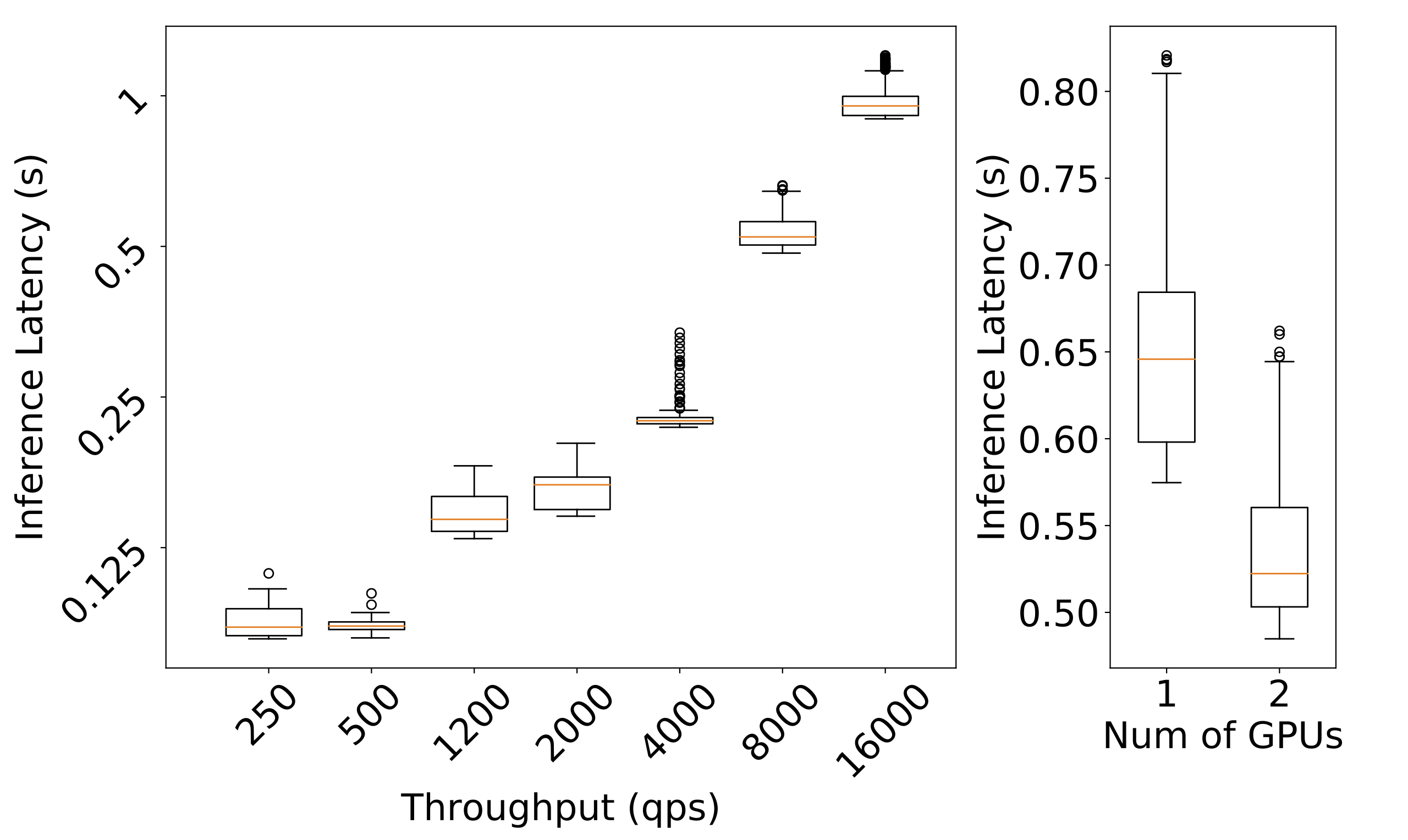}
    \vspace{-1em}
    \caption{
        Latency scalability. (Left): latency scales linearly with ingest rate from an increasing number of patients. (Right): latency improves with more GPUs. 
    }
\label{fig:boxplot_latency}
\vspace{-1.5em}
\end{figure}

In Figure \ref{fig:boxplot_latency} (left) we vary throughput by changing the number of patients and keeping the number of GPUs fixed (2 NVIDIA V100 GPUs). We generate a throughput of 250~qps from each patient in an open loop arrival process (i.e., not blocking on the results of prior queries). Throughput is then linear in the number of patients. Throughput increase contributes to (a) higher queueing delays and (b) more contention for the GPUs, increasing the end-to-end system response $T$ plotted on the y-axis.

As the number of patients increases, the throughput gets higher. Higher throughput leads to higher queueing delays which increases latency.  \mname serving system is able to perform inference of a 10 model ensemble within 1.15 seconds (95th percentile) even when the ingest throughput is high (64 patients * 250~qps = 16000 queries per second). 
In Figure \ref{fig:boxplot_latency} (right), we vary number of GPUs, keeping throughput fixed (16000 qps) to gauge its effect on latency. Due to less resource contention, ensemble served on two GPUs provides a lower latency compared to one GPU. 

\section{Conclusion}
\label{sec:conclusion}

ICU environment necessitates both timely and accurate decisions to manage patients' standard of care and cost of care. Clinical tasks, such as timely patient discharge, require expert medical staff. Deep learning models have achieved high accuracy of prediction on these tasks, but neglected latency implications. In this paper, we propose \mname---an online model ensemble serving system for DL models in ICU. \mname navigates the accuracy/latency tradeoff space, achieving the highest accuracy within specified sub-second latency targets. \mname's Ensemble Composer automatically constructs a model ensemble from a set of models trained for different sensory data modalities and, potentially, observation windows. \mname provides a real-time serving platform built on top of an open-source Ray~\cite{moritz2018ray} framework, with support for serving pipelines that consist of a combination of stateful (e.g., data aggregation) and stateless (e.g. model serving) components to achieve sub-second, latency SLO-aware performance. \mname is shown to outperform the conventional offline batched inference both on clinical prediction accuracy and latency (by order of magnitude).

\section{Acknowledgements}
This work was in part supported by the National Science Foundation award IIS-1418511, CCF-1533768 and IIS-1838042, the National Institute of Health award NIH R01 1R01NS107291-01 and R56HL138415.
\bibliographystyle{ACM-Reference-Format}
\bibliography{sample}


\begin{thebibliography}{39}


\ifx \showCODEN    \undefined \def \showCODEN     #1{\unskip}     \fi
\ifx \showDOI      \undefined \def \showDOI       #1{#1}\fi
\ifx \showISBNx    \undefined \def \showISBNx     #1{\unskip}     \fi
\ifx \showISBNxiii \undefined \def \showISBNxiii  #1{\unskip}     \fi
\ifx \showISSN     \undefined \def \showISSN      #1{\unskip}     \fi
\ifx \showLCCN     \undefined \def \showLCCN      #1{\unskip}     \fi
\ifx \shownote     \undefined \def \shownote      #1{#1}          \fi
\ifx \showarticletitle \undefined \def \showarticletitle #1{#1}   \fi
\ifx \showURL      \undefined \def \showURL       {\relax}        \fi
\providecommand\bibfield[2]{#2}
\providecommand\bibinfo[2]{#2}
\providecommand\natexlab[1]{#1}
\providecommand\showeprint[2][]{arXiv:#2}

\bibitem[\protect\citeauthoryear{Bailly, Meyfroidt, and Timsit}{Bailly
  et~al\mbox{.}}{2018}]%
        {bailly2018s}
\bibfield{author}{\bibinfo{person}{S{\'e}bastien Bailly},
  \bibinfo{person}{Geert Meyfroidt}, {and} \bibinfo{person}{Jean-Fran{\c{c}}ois
  Timsit}.} \bibinfo{year}{2018}\natexlab{}.
\newblock \showarticletitle{What’s new in ICU in 2050: big data and machine
  learning}.
\newblock \bibinfo{journal}{\emph{Intensive care medicine}}
  \bibinfo{volume}{44}, \bibinfo{number}{9} (\bibinfo{year}{2018}),
  \bibinfo{pages}{1524--1527}.
\newblock


\bibitem[\protect\citeauthoryear{Bergstra and Bengio}{Bergstra and
  Bengio}{2012}]%
        {bergstra2012random}
\bibfield{author}{\bibinfo{person}{James Bergstra} {and}
  \bibinfo{person}{Yoshua Bengio}.} \bibinfo{year}{2012}\natexlab{}.
\newblock \showarticletitle{Random search for hyper-parameter optimization}.
\newblock \bibinfo{journal}{\emph{Journal of machine learning research}}
  \bibinfo{volume}{13}, \bibinfo{number}{Feb} (\bibinfo{year}{2012}),
  \bibinfo{pages}{281--305}.
\newblock


\bibitem[\protect\citeauthoryear{Bergstra, Yamins, and Cox}{Bergstra
  et~al\mbox{.}}{2013}]%
        {bergstra2013making}
\bibfield{author}{\bibinfo{person}{James Bergstra}, \bibinfo{person}{Daniel
  Yamins}, {and} \bibinfo{person}{David~Daniel Cox}.}
  \bibinfo{year}{2013}\natexlab{}.
\newblock \showarticletitle{Making a science of model search: Hyperparameter
  optimization in hundreds of dimensions for vision architectures}.
\newblock \bibinfo{journal}{\emph{JMLR}} (\bibinfo{year}{2013}).
\newblock


\bibitem[\protect\citeauthoryear{Bergstra, Bardenet, Bengio, and
  K{\'e}gl}{Bergstra et~al\mbox{.}}{2011}]%
        {bergstra2011algorithms}
\bibfield{author}{\bibinfo{person}{James~S Bergstra}, \bibinfo{person}{R{\'e}mi
  Bardenet}, \bibinfo{person}{Yoshua Bengio}, {and} \bibinfo{person}{Bal{\'a}zs
  K{\'e}gl}.} \bibinfo{year}{2011}\natexlab{}.
\newblock \showarticletitle{Algorithms for hyper-parameter optimization}. In
  \bibinfo{booktitle}{\emph{Advances in neural information processing
  systems}}. \bibinfo{pages}{2546--2554}.
\newblock


\bibitem[\protect\citeauthoryear{Breiman}{Breiman}{1996}]%
        {bagging}
\bibfield{author}{\bibinfo{person}{Leo Breiman}.}
  \bibinfo{year}{1996}\natexlab{}.
\newblock \showarticletitle{Bagging Predictors}.
\newblock \bibinfo{journal}{\emph{Mach. Learn.}} \bibinfo{volume}{24},
  \bibinfo{number}{2} (\bibinfo{date}{Aug.} \bibinfo{year}{1996}),
  \bibinfo{pages}{123–140}.
\newblock
\showISSN{0885-6125}
\urldef\tempurl%
\url{https://doi.org/10.1023/A:1018054314350}
\showDOI{\tempurl}


\bibitem[\protect\citeauthoryear{Breiman}{Breiman}{2001}]%
        {breiman2001random}
\bibfield{author}{\bibinfo{person}{Leo Breiman}.}
  \bibinfo{year}{2001}\natexlab{}.
\newblock \showarticletitle{Random forests}.
\newblock \bibinfo{journal}{\emph{Machine learning}} \bibinfo{volume}{45},
  \bibinfo{number}{1} (\bibinfo{year}{2001}), \bibinfo{pages}{5--32}.
\newblock


\bibitem[\protect\citeauthoryear{Cai, Zhu, and Han}{Cai et~al\mbox{.}}{2018}]%
        {cai2018proxylessnas}
\bibfield{author}{\bibinfo{person}{Han Cai}, \bibinfo{person}{Ligeng Zhu},
  {and} \bibinfo{person}{Song Han}.} \bibinfo{year}{2018}\natexlab{}.
\newblock \showarticletitle{Proxylessnas: Direct neural architecture search on
  target task and hardware}.
\newblock \bibinfo{journal}{\emph{arXiv:1812.00332}} (\bibinfo{year}{2018}).
\newblock


\bibitem[\protect\citeauthoryear{Celi, Mark, Stone, and Montgomery}{Celi
  et~al\mbox{.}}{2013}]%
        {celi2013big}
\bibfield{author}{\bibinfo{person}{Leo~Anthony Celi}, \bibinfo{person}{Roger~G
  Mark}, \bibinfo{person}{David~J Stone}, {and} \bibinfo{person}{Robert~A
  Montgomery}.} \bibinfo{year}{2013}\natexlab{}.
\newblock \showarticletitle{“Big data” in the intensive care unit. Closing
  the data loop}.
\newblock \bibinfo{journal}{\emph{American journal of respiratory and critical
  care medicine}} \bibinfo{volume}{187}, \bibinfo{number}{11}
  (\bibinfo{year}{2013}), \bibinfo{pages}{1157}.
\newblock


\bibitem[\protect\citeauthoryear{Crankshaw, Wang, Zhou, Franklin, Gonzalez, and
  Stoica}{Crankshaw et~al\mbox{.}}{2016}]%
        {clipper-nsdi17}
\bibfield{author}{\bibinfo{person}{Daniel Crankshaw}, \bibinfo{person}{Xin
  Wang}, \bibinfo{person}{Giulio Zhou}, \bibinfo{person}{Michael~J. Franklin},
  \bibinfo{person}{Joseph~E. Gonzalez}, {and} \bibinfo{person}{Ion Stoica}.}
  \bibinfo{year}{2016}\natexlab{}.
\newblock \showarticletitle{Clipper: {A} Low-Latency Online Prediction Serving
  System}.
\newblock \bibinfo{journal}{\emph{CoRR}}  \bibinfo{volume}{abs/1612.03079}
  (\bibinfo{year}{2016}).
\newblock
\showeprint[arxiv]{1612.03079}
\urldef\tempurl%
\url{http://arxiv.org/abs/1612.03079}
\showURL{%
\tempurl}


\bibitem[\protect\citeauthoryear{Elsken, Metzen, and Hutter}{Elsken
  et~al\mbox{.}}{2018}]%
        {elsken2018neural}
\bibfield{author}{\bibinfo{person}{Thomas Elsken}, \bibinfo{person}{Jan~Hendrik
  Metzen}, {and} \bibinfo{person}{Frank Hutter}.}
  \bibinfo{year}{2018}\natexlab{}.
\newblock \showarticletitle{Neural architecture search: A survey}.
\newblock \bibinfo{journal}{\emph{arXiv preprint arXiv:1808.05377}}
  (\bibinfo{year}{2018}).
\newblock


\bibitem[\protect\citeauthoryear{Gershengorn, Garland, and Gong}{Gershengorn
  et~al\mbox{.}}{2015}]%
        {gershengorn2015patterns}
\bibfield{author}{\bibinfo{person}{Hayley~B Gershengorn},
  \bibinfo{person}{Allan Garland}, {and} \bibinfo{person}{Michelle~N Gong}.}
  \bibinfo{year}{2015}\natexlab{}.
\newblock \showarticletitle{Patterns of daily costs differ for medical and
  surgical intensive care unit patients}.
\newblock \bibinfo{journal}{\emph{Annals of the American Thoracic Society}}
  \bibinfo{volume}{12}, \bibinfo{number}{12} (\bibinfo{year}{2015}),
  \bibinfo{pages}{1831--1836}.
\newblock


\bibitem[\protect\citeauthoryear{Gulshan, Peng, Coram, Stumpe, Wu,
  Narayanaswamy, Venugopalan, Widner, Madams, Cuadros, Kim, Raman, Nelson,
  Mega, and Webster}{Gulshan et~al\mbox{.}}{2016}]%
        {google_jama}
\bibfield{author}{\bibinfo{person}{Varun Gulshan}, \bibinfo{person}{Lily Peng},
  \bibinfo{person}{Marc Coram}, \bibinfo{person}{Martin~C. Stumpe},
  \bibinfo{person}{Derek Wu}, \bibinfo{person}{Arunachalam Narayanaswamy},
  \bibinfo{person}{Subhashini Venugopalan}, \bibinfo{person}{Kasumi Widner},
  \bibinfo{person}{Tom Madams}, \bibinfo{person}{Jorge Cuadros},
  \bibinfo{person}{Ramasamy Kim}, \bibinfo{person}{Rajiv Raman},
  \bibinfo{person}{Philip~C. Nelson}, \bibinfo{person}{Jessica~L. Mega}, {and}
  \bibinfo{person}{Dale~R. Webster}.} \bibinfo{year}{2016}\natexlab{}.
\newblock \showarticletitle{{Development and Validation of a Deep Learning
  Algorithm for Detection of Diabetic Retinopathy in Retinal Fundus
  Photographs}}.
\newblock \bibinfo{journal}{\emph{JAMA}} \bibinfo{volume}{316},
  \bibinfo{number}{22} (\bibinfo{date}{12} \bibinfo{year}{2016}),
  \bibinfo{pages}{2402--2410}.
\newblock
\showISSN{0098-7484}
\urldef\tempurl%
\url{https://doi.org/10.1001/jama.2016.17216}
\showDOI{\tempurl}
\showeprint{https://jamanetwork.com/journals/jama/articlepdf/2588763/joi160132.pdf}


\bibitem[\protect\citeauthoryear{Halpern and Pastores}{Halpern and
  Pastores}{2015}]%
        {halpern2015critical}
\bibfield{author}{\bibinfo{person}{Neil~A Halpern} {and}
  \bibinfo{person}{Stephen~M Pastores}.} \bibinfo{year}{2015}\natexlab{}.
\newblock \showarticletitle{Critical care medicine beds, use, occupancy and
  costs in the United States: a methodological review}.
\newblock \bibinfo{journal}{\emph{Critical care medicine}}
  \bibinfo{volume}{43}, \bibinfo{number}{11} (\bibinfo{year}{2015}),
  \bibinfo{pages}{2452}.
\newblock


\bibitem[\protect\citeauthoryear{Hannun, Rajpurkar, Haghpanahi, Tison, Bourn,
  Turakhia, and Ng}{Hannun et~al\mbox{.}}{2019}]%
        {hannun2019cardiologist}
\bibfield{author}{\bibinfo{person}{Awni~Y Hannun}, \bibinfo{person}{Pranav
  Rajpurkar}, \bibinfo{person}{Masoumeh Haghpanahi},
  \bibinfo{person}{Geoffrey~H Tison}, \bibinfo{person}{Codie Bourn},
  \bibinfo{person}{Mintu~P Turakhia}, {and} \bibinfo{person}{Andrew~Y Ng}.}
  \bibinfo{year}{2019}\natexlab{}.
\newblock \showarticletitle{Cardiologist-level arrhythmia detection and
  classification in ambulatory electrocardiograms using a deep neural network}.
\newblock \bibinfo{journal}{\emph{Nature medicine}} \bibinfo{volume}{25},
  \bibinfo{number}{1} (\bibinfo{year}{2019}), \bibinfo{pages}{65}.
\newblock


\bibitem[\protect\citeauthoryear{Harutyunyan, Khachatrian, Kale, Steeg, and
  Galstyan}{Harutyunyan et~al\mbox{.}}{2017}]%
        {harutyunyan2017multitask}
\bibfield{author}{\bibinfo{person}{Hrayr Harutyunyan}, \bibinfo{person}{Hrant
  Khachatrian}, \bibinfo{person}{David~C Kale}, \bibinfo{person}{Greg~Ver
  Steeg}, {and} \bibinfo{person}{Aram Galstyan}.}
  \bibinfo{year}{2017}\natexlab{}.
\newblock \showarticletitle{Multitask learning and benchmarking with clinical
  time series data}.
\newblock \bibinfo{journal}{\emph{arXiv preprint arXiv:1703.07771}}
  (\bibinfo{year}{2017}).
\newblock


\bibitem[\protect\citeauthoryear{Hong, Zhou, Shang, Xiao, and Sun}{Hong
  et~al\mbox{.}}{2020}]%
        {hong2020opportunities}
\bibfield{author}{\bibinfo{person}{Shenda Hong}, \bibinfo{person}{Yuxi Zhou},
  \bibinfo{person}{Junyuan Shang}, \bibinfo{person}{Cao Xiao}, {and}
  \bibinfo{person}{Jimeng Sun}.} \bibinfo{year}{2020}\natexlab{}.
\newblock \showarticletitle{Opportunities and challenges of deep learning
  methods for electrocardiogram data: A systematic review}.
\newblock \bibinfo{journal}{\emph{Computers in Biology and Medicine}}
  (\bibinfo{year}{2020}), \bibinfo{pages}{103801}.
\newblock


\bibitem[\protect\citeauthoryear{Hutter, Hoos, and Leyton-Brown}{Hutter
  et~al\mbox{.}}{2011}]%
        {hutter2011sequential}
\bibfield{author}{\bibinfo{person}{Frank Hutter}, \bibinfo{person}{Holger~H
  Hoos}, {and} \bibinfo{person}{Kevin Leyton-Brown}.}
  \bibinfo{year}{2011}\natexlab{}.
\newblock \showarticletitle{Sequential model-based optimization for general
  algorithm configuration}. In \bibinfo{booktitle}{\emph{International
  conference on learning and intelligent optimization}}. Springer,
  \bibinfo{pages}{507--523}.
\newblock


\bibitem[\protect\citeauthoryear{Koziel and Leifsson}{Koziel and
  Leifsson}{2013}]%
        {koziel2013surrogate}
\bibfield{author}{\bibinfo{person}{Slawomir Koziel} {and}
  \bibinfo{person}{Leifur Leifsson}.} \bibinfo{year}{2013}\natexlab{}.
\newblock \bibinfo{booktitle}{\emph{Surrogate-based modeling and
  optimization}}.
\newblock \bibinfo{publisher}{Springer}.
\newblock


\bibitem[\protect\citeauthoryear{LeCun, Bengio, and Hinton}{LeCun
  et~al\mbox{.}}{2015}]%
        {lecun2015deep}
\bibfield{author}{\bibinfo{person}{Yann LeCun}, \bibinfo{person}{Yoshua
  Bengio}, {and} \bibinfo{person}{Geoffrey Hinton}.}
  \bibinfo{year}{2015}\natexlab{}.
\newblock \showarticletitle{Deep learning}.
\newblock \bibinfo{journal}{\emph{nature}} \bibinfo{volume}{521},
  \bibinfo{number}{7553} (\bibinfo{year}{2015}), \bibinfo{pages}{436}.
\newblock


\bibitem[\protect\citeauthoryear{L{\'e}vesque, Gagn{\'e}, and
  Sabourin}{L{\'e}vesque et~al\mbox{.}}{2016}]%
        {levesque2016bayesian}
\bibfield{author}{\bibinfo{person}{Julien-Charles L{\'e}vesque},
  \bibinfo{person}{Christian Gagn{\'e}}, {and} \bibinfo{person}{Robert
  Sabourin}.} \bibinfo{year}{2016}\natexlab{}.
\newblock \showarticletitle{Bayesian hyperparameter optimization for ensemble
  learning}. In \bibinfo{booktitle}{\emph{Proceedings of the Thirty-Second
  Conference on Uncertainty in Artificial Intelligence}}.
  \bibinfo{pages}{437--446}.
\newblock


\bibitem[\protect\citeauthoryear{Lipton, Kale, Elkan, and Wetzel}{Lipton
  et~al\mbox{.}}{2016}]%
        {DBLP:journals/corr/LiptonKEW15}
\bibfield{author}{\bibinfo{person}{Zachary~C Lipton}, \bibinfo{person}{David~C
  Kale}, \bibinfo{person}{Charles Elkan}, {and} \bibinfo{person}{Randall
  Wetzel}.} \bibinfo{year}{2016}\natexlab{}.
\newblock \showarticletitle{Learning to diagnose with LSTM recurrent neural
  networks}.
\newblock \bibinfo{journal}{\emph{ICLR}}.
\newblock


\bibitem[\protect\citeauthoryear{Liu, Zoph, Neumann, Shlens, Hua, Li, Fei-Fei,
  Yuille, Huang, and Murphy}{Liu et~al\mbox{.}}{2018}]%
        {liu2018progressive}
\bibfield{author}{\bibinfo{person}{Chenxi Liu}, \bibinfo{person}{Barret Zoph},
  \bibinfo{person}{Maxim Neumann}, \bibinfo{person}{Jonathon Shlens},
  \bibinfo{person}{Wei Hua}, \bibinfo{person}{Li-Jia Li}, \bibinfo{person}{Li
  Fei-Fei}, \bibinfo{person}{Alan Yuille}, \bibinfo{person}{Jonathan Huang},
  {and} \bibinfo{person}{Kevin Murphy}.} \bibinfo{year}{2018}\natexlab{}.
\newblock \showarticletitle{Progressive neural architecture search}. In
  \bibinfo{booktitle}{\emph{ECCV}}. \bibinfo{pages}{19--34}.
\newblock


\bibitem[\protect\citeauthoryear{Meng, Bradley, Yavuz, Sparks, Venkataraman,
  Liu, Freeman, Tsai, Amde, Owen, Xin, Xin, Franklin, Zadeh, Zaharia, and
  Talwalkar}{Meng et~al\mbox{.}}{2015}]%
        {meng2015mllib}
\bibfield{author}{\bibinfo{person}{Xiangrui Meng}, \bibinfo{person}{Joseph
  Bradley}, \bibinfo{person}{Burak Yavuz}, \bibinfo{person}{Evan Sparks},
  \bibinfo{person}{Shivaram Venkataraman}, \bibinfo{person}{Davies Liu},
  \bibinfo{person}{Jeremy Freeman}, \bibinfo{person}{DB Tsai},
  \bibinfo{person}{Manish Amde}, \bibinfo{person}{Sean Owen},
  \bibinfo{person}{Doris Xin}, \bibinfo{person}{Reynold Xin},
  \bibinfo{person}{Michael~J. Franklin}, \bibinfo{person}{Reza Zadeh},
  \bibinfo{person}{Matei Zaharia}, {and} \bibinfo{person}{Ameet Talwalkar}.}
  \bibinfo{year}{2015}\natexlab{}.
\newblock \bibinfo{title}{MLlib: Machine Learning in Apache Spark}.
\newblock
\newblock
\showeprint[arxiv]{1505.06807}~[cs.LG]


\bibitem[\protect\citeauthoryear{Mockus}{Mockus}{2012}]%
        {mockus2012bayesian}
\bibfield{author}{\bibinfo{person}{Jonas Mockus}.}
  \bibinfo{year}{2012}\natexlab{}.
\newblock \bibinfo{booktitle}{\emph{Bayesian approach to global optimization:
  theory and applications}}. Vol.~\bibinfo{volume}{37}.
\newblock \bibinfo{publisher}{Springer Science \& Business Media}.
\newblock


\bibitem[\protect\citeauthoryear{Moritz, Nishihara, Wang, Tumanov, Liaw, Liang,
  Elibol, Yang, Paul, Jordan, et~al\mbox{.}}{Moritz et~al\mbox{.}}{2018}]%
        {moritz2018ray}
\bibfield{author}{\bibinfo{person}{Philipp Moritz}, \bibinfo{person}{Robert
  Nishihara}, \bibinfo{person}{Stephanie Wang}, \bibinfo{person}{Alexey
  Tumanov}, \bibinfo{person}{Richard Liaw}, \bibinfo{person}{Eric Liang},
  \bibinfo{person}{Melih Elibol}, \bibinfo{person}{Zongheng Yang},
  \bibinfo{person}{William Paul}, \bibinfo{person}{Michael~I Jordan},
  {et~al\mbox{.}}} \bibinfo{year}{2018}\natexlab{}.
\newblock \showarticletitle{Ray: A distributed framework for emerging
  $\{$AI$\}$ applications}. In \bibinfo{booktitle}{\emph{13th $\{$USENIX$\}$
  Symposium on Operating Systems Design and Implementation ($\{$OSDI$\}$ 18)}}.
  \bibinfo{pages}{561--577}.
\newblock


\bibitem[\protect\citeauthoryear{Nguyen, Tran, and Venkatesh}{Nguyen
  et~al\mbox{.}}{2017}]%
        {nguyen2017deep}
\bibfield{author}{\bibinfo{person}{Phuoc Nguyen}, \bibinfo{person}{Truyen
  Tran}, {and} \bibinfo{person}{Svetha Venkatesh}.}
  \bibinfo{year}{2017}\natexlab{}.
\newblock \showarticletitle{Deep learning to attend to risk in ICU}.
\newblock \bibinfo{journal}{\emph{arXiv preprint arXiv:1707.05010}}
  (\bibinfo{year}{2017}).
\newblock


\bibitem[\protect\citeauthoryear{Olston, Fiedel, Gorovoy, Harmsen, Lao, Li,
  Rajashekhar, Ramesh, and Soyke}{Olston et~al\mbox{.}}{2017}]%
        {olston2017tensorflow}
\bibfield{author}{\bibinfo{person}{Christopher Olston}, \bibinfo{person}{Noah
  Fiedel}, \bibinfo{person}{Kiril Gorovoy}, \bibinfo{person}{Jeremiah Harmsen},
  \bibinfo{person}{Li Lao}, \bibinfo{person}{Fangwei Li}, \bibinfo{person}{Vinu
  Rajashekhar}, \bibinfo{person}{Sukriti Ramesh}, {and} \bibinfo{person}{Jordan
  Soyke}.} \bibinfo{year}{2017}\natexlab{}.
\newblock \showarticletitle{Tensorflow-serving: Flexible, high-performance ml
  serving}.
\newblock \bibinfo{journal}{\emph{arXiv:1712.06139}} (\bibinfo{year}{2017}).
\newblock


\bibitem[\protect\citeauthoryear{Paszke, Gross, Massa, Lerer, Bradbury, Chanan,
  Killeen, Lin, Gimelshein, Antiga, et~al\mbox{.}}{Paszke
  et~al\mbox{.}}{2019}]%
        {paszke2019pytorch}
\bibfield{author}{\bibinfo{person}{Adam Paszke}, \bibinfo{person}{Sam Gross},
  \bibinfo{person}{Francisco Massa}, \bibinfo{person}{Adam Lerer},
  \bibinfo{person}{James Bradbury}, \bibinfo{person}{Gregory Chanan},
  \bibinfo{person}{Trevor Killeen}, \bibinfo{person}{Zeming Lin},
  \bibinfo{person}{Natalia Gimelshein}, \bibinfo{person}{Luca Antiga},
  {et~al\mbox{.}}} \bibinfo{year}{2019}\natexlab{}.
\newblock \showarticletitle{PyTorch: An imperative style, high-performance deep
  learning library}. In \bibinfo{booktitle}{\emph{Advances in Neural
  Information Processing Systems}}. \bibinfo{pages}{8024--8035}.
\newblock


\bibitem[\protect\citeauthoryear{Rajkomar, Oren, Chen, Dai, Hajaj, Hardt, Liu,
  Liu, Marcus, Sun, et~al\mbox{.}}{Rajkomar et~al\mbox{.}}{2018}]%
        {rajkomar2018scalable}
\bibfield{author}{\bibinfo{person}{Alvin Rajkomar}, \bibinfo{person}{Eyal
  Oren}, \bibinfo{person}{Kai Chen}, \bibinfo{person}{Andrew~M Dai},
  \bibinfo{person}{Nissan Hajaj}, \bibinfo{person}{Michaela Hardt},
  \bibinfo{person}{Peter~J Liu}, \bibinfo{person}{Xiaobing Liu},
  \bibinfo{person}{Jake Marcus}, \bibinfo{person}{Mimi Sun}, {et~al\mbox{.}}}
  \bibinfo{year}{2018}\natexlab{}.
\newblock \showarticletitle{Scalable and accurate deep learning with electronic
  health records}.
\newblock \bibinfo{journal}{\emph{NPJ Digital Medicine}} \bibinfo{volume}{1},
  \bibinfo{number}{1} (\bibinfo{year}{2018}), \bibinfo{pages}{18}.
\newblock


\bibitem[\protect\citeauthoryear{Rajpurkar, Irvin, Zhu, Yang, Mehta, Duan,
  Ding, Bagul, Langlotz, Shpanskaya, et~al\mbox{.}}{Rajpurkar
  et~al\mbox{.}}{2017}]%
        {rajpurkar2017chexnet}
\bibfield{author}{\bibinfo{person}{Pranav Rajpurkar}, \bibinfo{person}{Jeremy
  Irvin}, \bibinfo{person}{Kaylie Zhu}, \bibinfo{person}{Brandon Yang},
  \bibinfo{person}{Hershel Mehta}, \bibinfo{person}{Tony Duan},
  \bibinfo{person}{Daisy Ding}, \bibinfo{person}{Aarti Bagul},
  \bibinfo{person}{Curtis Langlotz}, \bibinfo{person}{Katie Shpanskaya},
  {et~al\mbox{.}}} \bibinfo{year}{2017}\natexlab{}.
\newblock \showarticletitle{Chexnet: Radiologist-level pneumonia detection on
  chest x-rays with deep learning}.
\newblock \bibinfo{journal}{\emph{arXiv preprint arXiv:1711.05225}}
  (\bibinfo{year}{2017}).
\newblock


\bibitem[\protect\citeauthoryear{Shahriari, Swersky, Wang, Adams, and
  De~Freitas}{Shahriari et~al\mbox{.}}{2015}]%
        {shahriari2015taking}
\bibfield{author}{\bibinfo{person}{Bobak Shahriari}, \bibinfo{person}{Kevin
  Swersky}, \bibinfo{person}{Ziyu Wang}, \bibinfo{person}{Ryan~P Adams}, {and}
  \bibinfo{person}{Nando De~Freitas}.} \bibinfo{year}{2015}\natexlab{}.
\newblock \showarticletitle{Taking the human out of the loop: A review of
  Bayesian optimization}.
\newblock \bibinfo{journal}{\emph{Proc. IEEE}} \bibinfo{volume}{104},
  \bibinfo{number}{1} (\bibinfo{year}{2015}), \bibinfo{pages}{148--175}.
\newblock


\bibitem[\protect\citeauthoryear{Snoek, Larochelle, and Adams}{Snoek
  et~al\mbox{.}}{2012}]%
        {snoek2012practical}
\bibfield{author}{\bibinfo{person}{Jasper Snoek}, \bibinfo{person}{Hugo
  Larochelle}, {and} \bibinfo{person}{Ryan~P Adams}.}
  \bibinfo{year}{2012}\natexlab{}.
\newblock \showarticletitle{Practical bayesian optimization of machine learning
  algorithms}. In \bibinfo{booktitle}{\emph{Advances in neural information
  processing systems}}. \bibinfo{pages}{2951--2959}.
\newblock


\bibitem[\protect\citeauthoryear{Wang, Zhang, He, and Zha}{Wang
  et~al\mbox{.}}{2018}]%
        {wang2018supervised}
\bibfield{author}{\bibinfo{person}{Lu Wang}, \bibinfo{person}{Wei Zhang},
  \bibinfo{person}{Xiaofeng He}, {and} \bibinfo{person}{Hongyuan Zha}.}
  \bibinfo{year}{2018}\natexlab{}.
\newblock \showarticletitle{Supervised reinforcement learning with recurrent
  neural network for dynamic treatment recommendation}. In
  \bibinfo{booktitle}{\emph{Proceedings of the 24th ACM SIGKDD International
  Conference on Knowledge Discovery \& Data Mining}}. ACM,
  \bibinfo{pages}{2447--2456}.
\newblock


\bibitem[\protect\citeauthoryear{Whitley}{Whitley}{1994}]%
        {whitley1994genetic}
\bibfield{author}{\bibinfo{person}{Darrell Whitley}.}
  \bibinfo{year}{1994}\natexlab{}.
\newblock \showarticletitle{A genetic algorithm tutorial}.
\newblock \bibinfo{journal}{\emph{Statistics and computing}}
  \bibinfo{volume}{4}, \bibinfo{number}{2} (\bibinfo{year}{1994}),
  \bibinfo{pages}{65--85}.
\newblock


\bibitem[\protect\citeauthoryear{Xiao, Choi, and Sun}{Xiao
  et~al\mbox{.}}{2018}]%
        {xiao2018opportunities}
\bibfield{author}{\bibinfo{person}{Cao Xiao}, \bibinfo{person}{Edward Choi},
  {and} \bibinfo{person}{Jimeng Sun}.} \bibinfo{year}{2018}\natexlab{}.
\newblock \showarticletitle{Opportunities and challenges in developing deep
  learning models using electronic health records data: a systematic review}.
\newblock \bibinfo{journal}{\emph{JAMIA}} \bibinfo{volume}{25},
  \bibinfo{number}{10} (\bibinfo{year}{2018}), \bibinfo{pages}{1419--1428}.
\newblock


\bibitem[\protect\citeauthoryear{Xie, Girshick, Doll{\'a}r, Tu, and He}{Xie
  et~al\mbox{.}}{2017}]%
        {xie2017aggregated}
\bibfield{author}{\bibinfo{person}{Saining Xie}, \bibinfo{person}{Ross
  Girshick}, \bibinfo{person}{Piotr Doll{\'a}r}, \bibinfo{person}{Zhuowen Tu},
  {and} \bibinfo{person}{Kaiming He}.} \bibinfo{year}{2017}\natexlab{}.
\newblock \showarticletitle{Aggregated residual transformations for deep neural
  networks}. In \bibinfo{booktitle}{\emph{Proceedings of the IEEE conference on
  computer vision and pattern recognition}}. \bibinfo{pages}{1492--1500}.
\newblock


\bibitem[\protect\citeauthoryear{Xu, Biswal, Deshpande, Maher, and Sun}{Xu
  et~al\mbox{.}}{2018}]%
        {xu2018raim}
\bibfield{author}{\bibinfo{person}{Yanbo Xu}, \bibinfo{person}{Siddharth
  Biswal}, \bibinfo{person}{Shriprasad~R Deshpande}, \bibinfo{person}{Kevin~O
  Maher}, {and} \bibinfo{person}{Jimeng Sun}.} \bibinfo{year}{2018}\natexlab{}.
\newblock \showarticletitle{Raim: Recurrent attentive and intensive model of
  multimodal patient monitoring data}. In \bibinfo{booktitle}{\emph{KDD}}. ACM,
  \bibinfo{pages}{2565--2573}.
\newblock


\bibitem[\protect\citeauthoryear{Zhou}{Zhou}{2012}]%
        {zhou2012ensemble}
\bibfield{author}{\bibinfo{person}{Zhi-Hua Zhou}.}
  \bibinfo{year}{2012}\natexlab{}.
\newblock \bibinfo{booktitle}{\emph{Ensemble methods: foundations and
  algorithms}}.
\newblock \bibinfo{publisher}{Chapman and Hall/CRC}.
\newblock


\bibitem[\protect\citeauthoryear{Zoph and Le}{Zoph and Le}{2016}]%
        {zoph2016neural}
\bibfield{author}{\bibinfo{person}{Barret Zoph} {and} \bibinfo{person}{Quoc~V
  Le}.} \bibinfo{year}{2016}\natexlab{}.
\newblock \showarticletitle{Neural architecture search with reinforcement
  learning}.
\newblock \bibinfo{journal}{\emph{arXiv preprint arXiv:1611.01578}}
  (\bibinfo{year}{2016}).
\newblock


\end{thebibliography}

\clearpage
\section*{Supplementary materials}
\noindent \textbf{A.1 Exploring in \mname using genetic search}

\begin{algorithm}[H]
\caption{ Explore using Genetic algorithm }
\label{alg:explore}
\begin{algorithmic}[1]
\STATE Input: $\bm{B}$, number of samples $N_1$, degree of mutation $S$, probability of genetic explore $p$, probability of mutation $p_1$.
\STATE Output: $\bm{B}'$
\WHILE {i < $N_1$}
\STATE Uniformly random pick two numbers $rnd$, $rnd_1$ from [0,1]
\STATE Random pick $\bm{b}_1$, $\bm{b}_2$, $\bm{b}_3$ from $\bm{B}$
\IF {$rnd$ > $p$}
\STATE \textcolor{blue}{\texttt{/* Random explore */}}
\STATE $\bm{b}$ = Random($\mathcal{B}$)
\ELSE {}
\IF {$rnd_1$ > $p_1$}
\STATE \textcolor{blue}{\texttt{/* Recombination explore */}}
\STATE $\bm{b}$ = Recombination($\bm{b}_1$, $\bm{b}_2$)
\ELSE {}
\STATE \textcolor{blue}{\texttt{/* Mutation explore */}}
\STATE $\bm{b}$ = Mutation($\bm{b}_3$, $S$)
\ENDIF
\ENDIF
\STATE \textcolor{blue}{\texttt{/* Not add duplicates */}}
\IF {$\bm{b} \in \bm{B}$ or $\bm{b} \in \bm{B}'$}
\STATE \textbf{continue}
\ELSE {}
\STATE i = i + 1
\STATE \textcolor{blue}{\texttt{/* Add it to candidates */}}
\STATE $\bm{B}'=\bm{B}' \bigcup \bm{b}$
\ENDIF
\ENDWHILE
\end{algorithmic}
\end{algorithm}

\noindent \textbf{A.2 Evolution of ROC-AUC and latency when exploring the searching space using different algorithms}

\noindent \textbf{A.3 Comparison of accuracy and latency from the optimal ensemble selected by different algorithms}
\begin{figure}[H]
\raggedleft
\includegraphics[width=\linewidth]{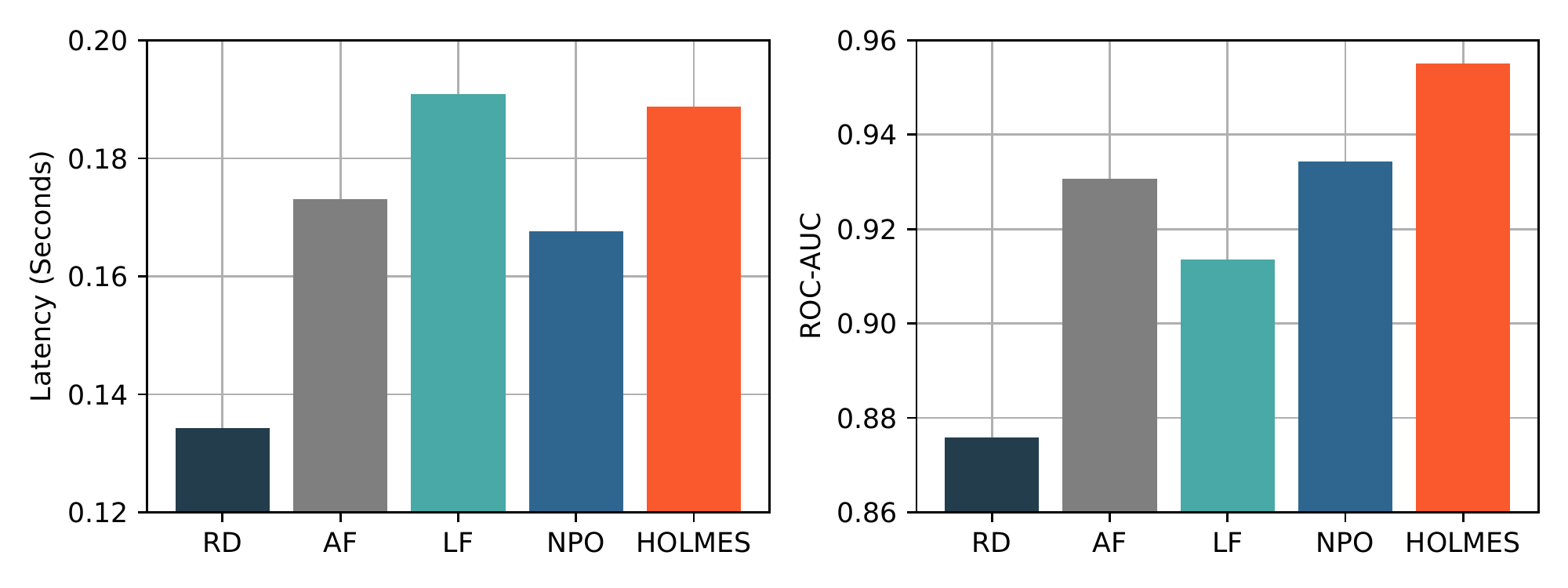}
\caption{ Comparing under 0.2 second latency constraint. (Left): \mname has comparable utility of latency with LF (Latency First) method. (Right): \mname selects higher accurate ensemble model under that latency constraint.
}
\label{fig:intro_app}
\end{figure}

\begin{figure}[ht]
\begin{tabular}{cc}
\includegraphics[width=0.45\linewidth]{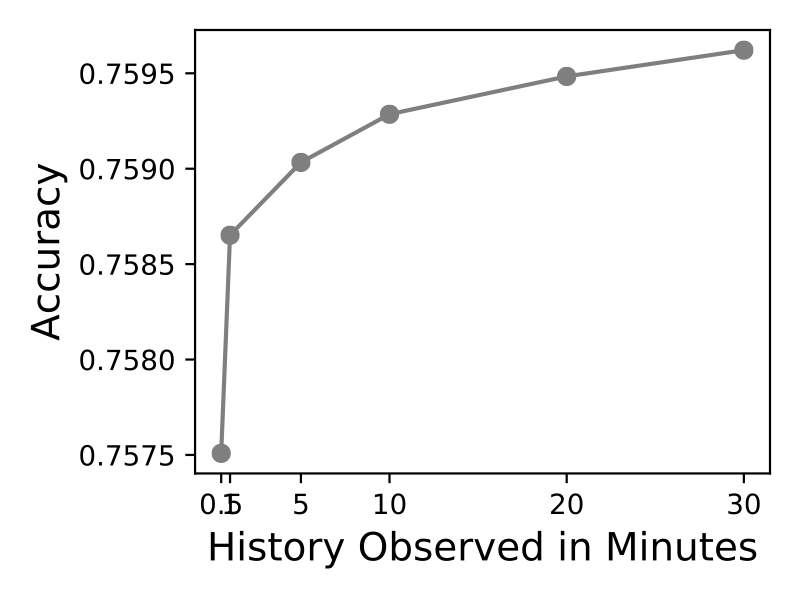} 
&  \includegraphics[width=0.45\linewidth]{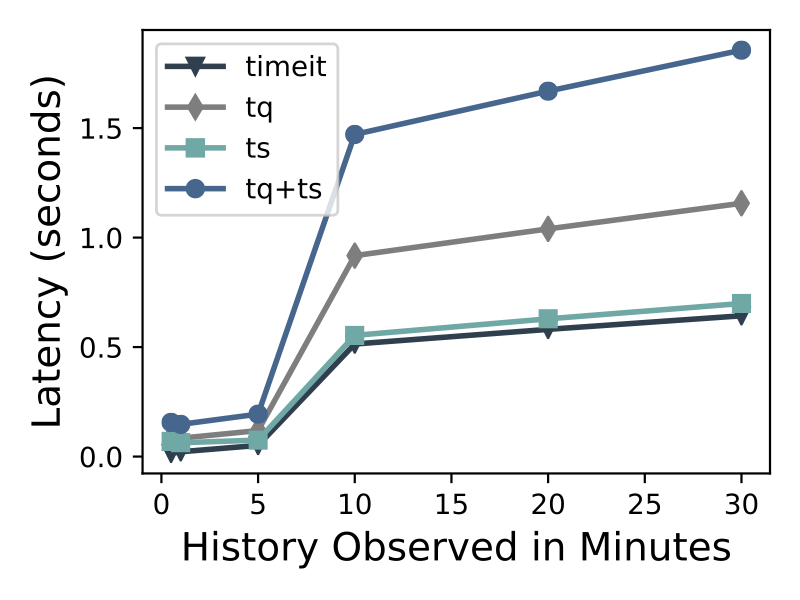} \\
\end{tabular}{}
\caption{Effects of different history aggregation.}
\label{fig:history}
\end{figure}

\noindent \textbf{A.4 Change of accuracy and latency by varying the observation window.}
In Figure \ref{fig:history}, we do comparison of execution in latency profile to show how does increase in observation window causes a small increase in latency while leads to high latency profile. Our right graph shows multiple execution of a model inside HOLMES serving system. Timeit (Time in PyTorch) legend shows the execution of plain PyTorch model in a GPU. TS (serving delay) legend shows the execution of the model inside the serving system. TQ (queuing delay) legend  
shows the worst analysis of a request waiting in a queue inside the serving system. TQ + TS legend shows end-to-end latency for a query inside the serving system.

\noindent \textbf{A.5 Deep model description in the Model Zoo.  }

\begin{table}[ht]
\caption{Deep model description in the Model Zoo. }
\centering
\begin{tabular}{l|l}
\toprule
\textbf{Field} & \textbf{Description}  \\
\midrule
Depth & Number of stacked layers \\ 
Width & Number of convolutional filters \\ 
MACS & Multiply-accumulate operations \\
Memory size & GPU memory usage \\
Input data modality & ECG Lead number or vital sign names\\
Input data length & Length of each input signal segmentation \\ 
Accuracy & ROC-AUC on validation set  \\
\bottomrule
\end{tabular}
\label{tb:model_profile}
\end{table}

\noindent \textbf{A.6 An Alternative Formulation of Accuracy Sensitive Constraint.  }

In the main paper, as desired for a real-time serving system, we have formulated our problem as a latency sensitive task that aims to maximize accuracy subject to a latency upper bound $L$. Alternatively, for other accuracy sensitive tasks, we can switch the objective function and constraint in Eq. (\ref{eq:latency}) and reach at a new Lagrange function:
\begin{equation*}
\begin{aligned}
\min_{\bm{b} \in \{0,1\}^{n}} \quad & L_l(\bm{b}) = f_l(\bm{V}, \bm{c}, \bm{b}) - \delta(f_a(\bm{V}, \bm{b})-A),\\
\end{aligned}\label{eq:opt2}
\end{equation*}
where $A$ is the accuracy lower bound that a model ensemble needs to achieve at least. Although this setup is beyond the scope of this paper, it can be equivalently solved by following the same searching algorithm we propose in the following subsection. 
\end{document}